\documentclass[letterpaper, 10pt, conference]{ieeeconf}
\IEEEoverridecommandlockouts % Only needed if you want to use the \thanks command
\usepackage{graphicx}
\usepackage{amsmath} % assumes amsmath package installed
\usepackage{amssymb}  % assumes amsmath package installed
\usepackage{hhline}
\usepackage{listings}
\usepackage{subfig}
\usepackage{soul}
\usepackage{hyperref}
\usepackage{multirow}
\usepackage[ruled]{algorithm2e}

\graphicspath{ {figures/} }

\newlength{\tempdima}
\newcommand{\rowname}[1]% #1 = text
{\rotatebox{90}{\makebox[\tempdima][c]{#1}}}

\newcommand{\red}[1]% #1 = text
{\colorbox{red!30}{#1}}

\usepackage{xcolor}
\definecolor{MyOrange}{RGB}{255, 109, 49} %// 255, 109, 49 | 255, 114, 96 %http://www.colorcombos.com/color-schemes/285/ColorCombo285.html
\definecolor{MyGreen}{RGB}{18, 151, 147}
\definecolor{MyGray}{RGB}{30, 30, 30}
\definecolor{MyBlue}{RGB}{155, 215, 213}
\definecolor{MyCream}{RGB}{255, 245, 195}
\definecolor{MyYellow}{RGB}{255, 203, 24}
\definecolor{MyBrown}{RGB}{187, 119, 46}
\definecolor{MyPurple}{RGB}{111, 54, 98}
\definecolor{CommentGreen}{HTML}{008000}
\definecolor{StringRed}{HTML}{A31515}

\title{\LARGE \bf
Co-Fusion: Real-time  Segmentation, Tracking and Fusion of Multiple Objects}

\author{Martin R{\"u}nz and Lourdes Agapito\\% <-this % stops a space
Department of Computer Science, University College London, UK\\
       {\tt\small{\{martin.runz.15,l.agapito\}@ucl.ac.uk}}\\
{\tt\small{\url{http://visual.cs.ucl.ac.uk/pubs/cofusion/index.html}}}%
}

\begin{document}

%\maketitle
\thispagestyle{empty}
\pagestyle{empty}

\newcommand{\bD}{\boldsymbol{D}}
\newcommand{\R}{\mathbb{R}}
\newcommand{\bS}{\boldsymbol{S}}
\newcommand{\bA}{\boldsymbol{A}}
\newcommand{\bs}{\boldsymbol{s}}
\newcommand{\bT}{\boldsymbol{T}}
\newcommand{\bX}{\boldsymbol{X}}
\newcommand{\bbarS}{\boldsymbol{\bar{S}}}
\newcommand{\bR}{\boldsymbol{R}}
\newcommand{\bq}{\boldsymbol{q}}
\newcommand{\bW}{\boldsymbol{W}}
\newcommand{\bV}{\boldsymbol{V}}
\newcommand{\bL}{\boldsymbol{L}}
\newcommand{\bt}{\boldsymbol{t}}
\newcommand{\bx}{\boldsymbol{x}}
\newcommand{\bu}{\boldsymbol{u}}
\newcommand{\bI}{\boldsymbol{I}}
\newcommand{\nrsfm}[0]{{\sc nrs{\rm\em f}\sc m}}
\newcommand{\sfm}[0]{{\sc s{\rm\em f}\sc m}}
\newcommand{\gpu}[0]{{\sc gpu}}
\newcommand{\mvs}[0]{{\sc mvs}}
\newcommand{\xmark}{\ding{55}}

%%%%%%%%%%%%%%%%%%%%%%%%%%%%%%%%%%%%%%%%%%%%%%%%%%%%%%%%%%%%%%%%%%%%%%%%%%%%%%%%

\twocolumn[{
\renewcommand\twocolumn[1][]{#1}
\maketitle
\begin{center}
    \centering
      \includegraphics[width=2.0\columnwidth]{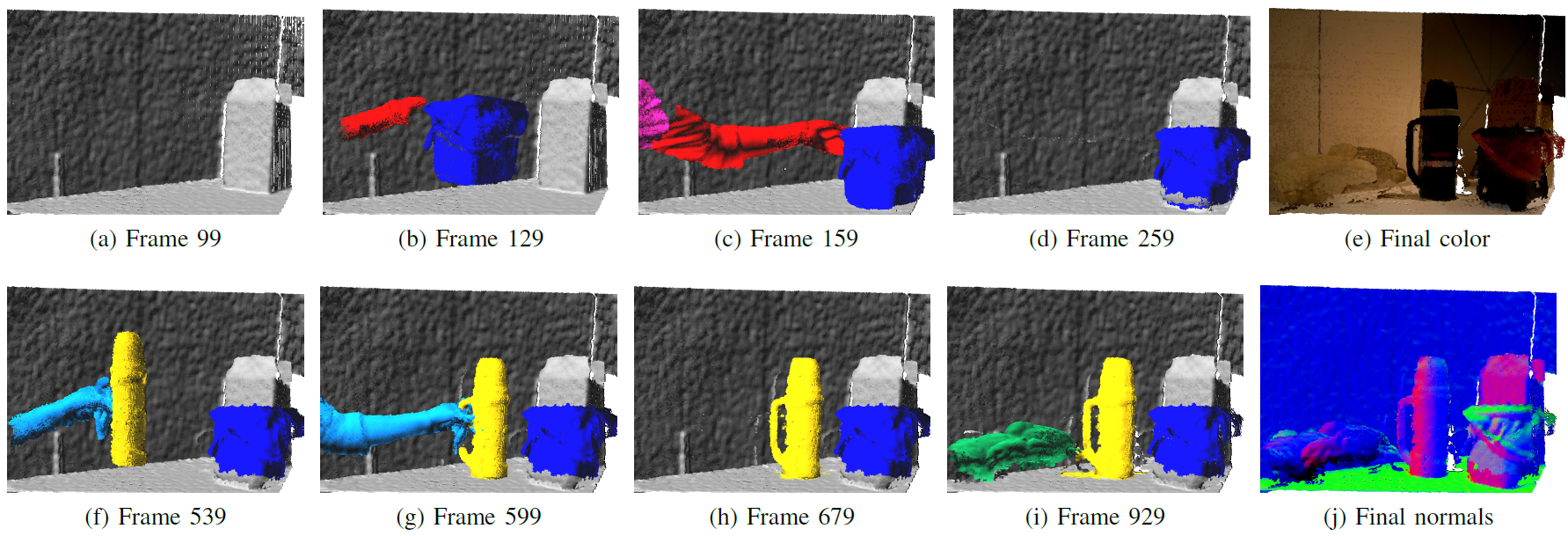} \captionof{figure}{A sequence demonstrating our dynamic SLAM system. Three objects
    were sequentially placed on a table: first a small bin (blue label),
    a flask (yellow) and a teddy bear (green). The results show that all objects were successfully segmented, tracked and modeled.}
\label{fig:placed-items}
\end{center}%
}]

%%%%%%%%%%%%%%%%%%%%%%%%%%%%%%%%%%%%%%%%%%%%%%%%%%%%%%%%%%%%%%%%%%%%%%%%%%%%%%%%

\begin{abstract}

In this paper we introduce Co-Fusion, a dense SLAM system that
takes a live stream of RGB-D images as input and segments the scene
into different objects (using either motion or semantic cues) while
simultaneously tracking and reconstructing their 3D shape in real
time.
We use a multiple model fitting approach where each object can move
independently from the background and still be effectively tracked and
its shape fused over time using only the information from pixels
associated with that object label. Previous attempts to deal with
dynamic scenes have typically considered moving regions as outliers,
and consequently do not model their shape or track their motion over
time. In contrast, we enable the robot to maintain 3D models for each
of the segmented objects and to improve them over time through
fusion. As a result, our system can enable a robot to
maintain a scene description at the object level which has the
potential to allow interactions with its working environment; even in
the case of dynamic scenes.

\end{abstract}

%%%%%%%%%%%%%%%%%%%%%%%%%%%%%%%%%%%%%%%%%%%%%%%%%%%%%%%%%%%%%%%%%%%%%%%%%%%%%%%%
\section{INTRODUCTION}

The wide availability of affordable structured light and time of flight
depth sensors has had enormous impact both on the democratization of
the acquisition of 3D models in real time from hand-held cameras and
on providing robots with powerful but low-cost 3D sensing
capabilities. Tracking the motion of a camera while maintaining a
dense representation of the 3D geometry of its environment in real
time has become more important than
ever~\cite{kinect-fusion,kintinuous,elastic-fusion,dynamic-fusion}.

While solid progress has been made towards solving this problem in the
case of static environments, where the only motion is that of the
camera, dealing with dynamic scenes where an unknown number of objects
might be moving independently is significantly harder. The typical
strategy adopted by most systems is to track only the motion of the
camera relative to the static background and treat moving objects as
outliers whose 3D geometry and motion is not modeled over
time. However, in robotics applications often it is precisely the
objects moving in the foreground that are of most interest to the
robot. If we want to design robots that can interact with dynamic
scenes it is crucial to equip them with the capability to \emph{(i)}
discover objects in the scene via segmentation \emph{(ii)} track and
estimate the 3D geometry of each object independently. These high
level \emph{object-based} representations of the scene would greatly
enhance the perception and physical interaction capabilities of a
robot.

Consider for instance a SLAM system on-board a self-driving car
-- tracking and maintaining 3D models of all the moving cars around it
and not just the static parts of the scene
could be critical to avoid collisions.
Or think of a robot that arrives at a scene without a priori 3D
knowledge about the objects it must interact with -- the ability to
segment, track and fuse different objects would allow it actively to
discover and learn accurate 3D models of them on the fly through
motion, by picking them up, pushing them around or simply observing
how they move. An \emph{object level} scene description of this kind,
has the potential to enable the robot to interact physically with
the scene.

In this paper we introduce Co-FUSION a new RGB-D based SLAM system
that can segment a scene into the background and different foreground
objects, using either motion or semantic cues, while simultaneously
tracking and reconstructing their 3D geometry over time. Our
underlying assumption is that objects of interest can be detected and
segmented in real-time using efficient segmentation algorithms and
then tracked independently over time. Our system offers two
alternative grouping strategies -- \emph{motion segmentation} that
groups together points that move consistently in 3D and \emph{object
instance segmentation} that both detects and segments individual
objects of interest (at the pixel level) in an RGB image given a
semantic label. These two forms of segmentation allow us not only to
detect objects due to their motion but also objects that might be
static but are semantically of interest to the robot.

\begin{figure*}[ht]
\center
\def \imwidth {.21}
\settoheight{\tempdima}{\includegraphics[width=\imwidth\linewidth]{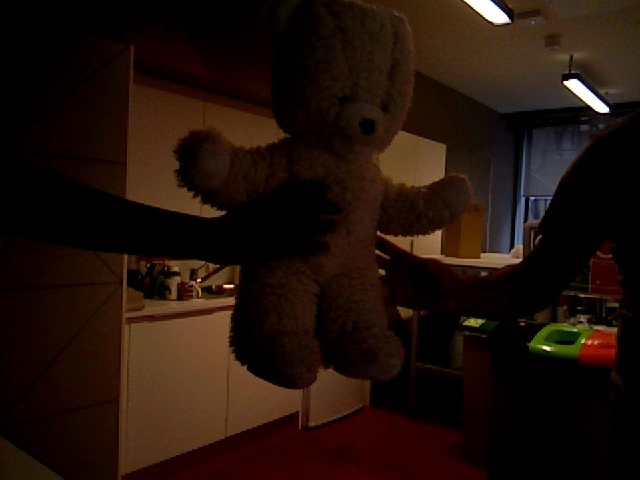}}
\begin{tabular}{@{}c@{ }c@{ }c@{ }c@{ }c@{}}

& RGB & Depth & Segmentation & 3D \\

\rowname{Before (frame 300)} &
\includegraphics[width=\imwidth\linewidth]{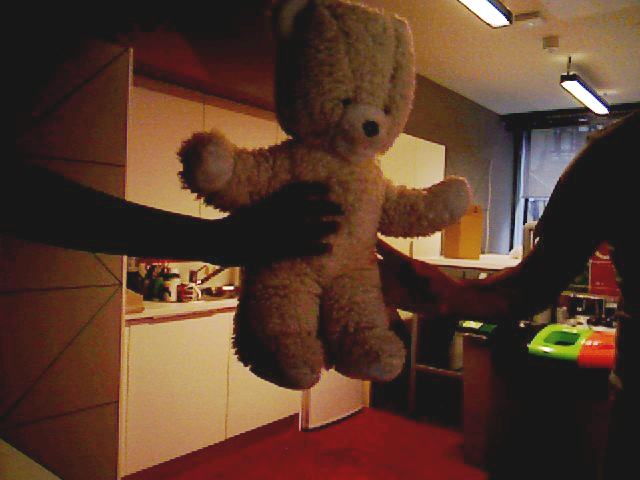} &
\includegraphics[width=\imwidth\linewidth]{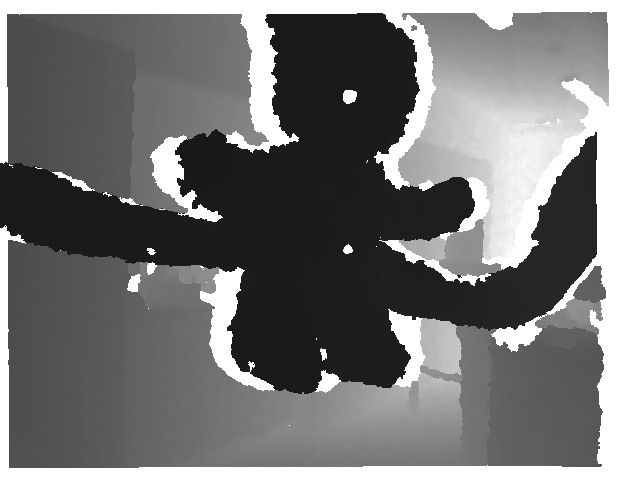} &
\includegraphics[width=\imwidth\linewidth]{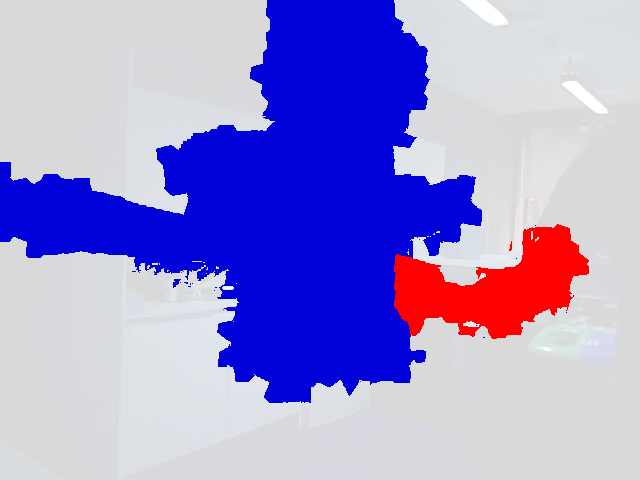} &
\includegraphics[width=\imwidth\linewidth]{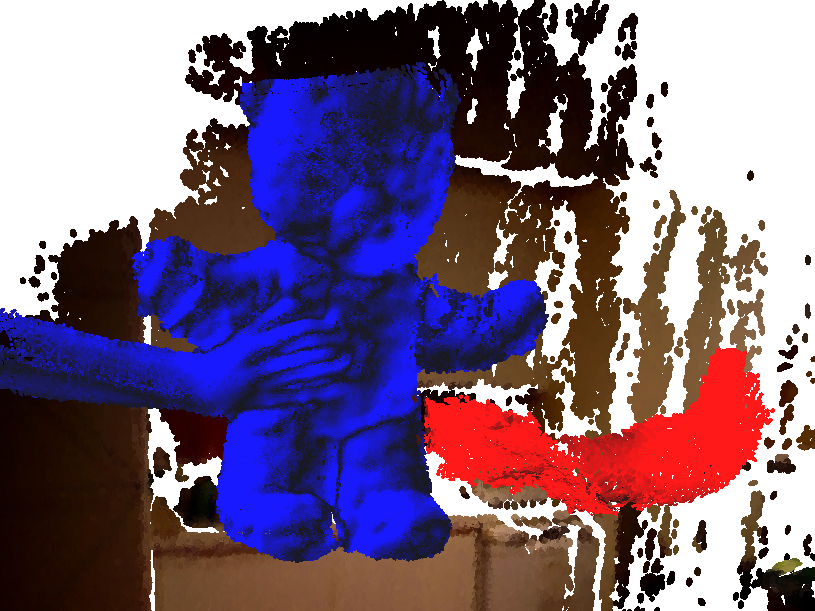} \\

\rowname{After (frame 370)} &
\includegraphics[width=\imwidth\linewidth]{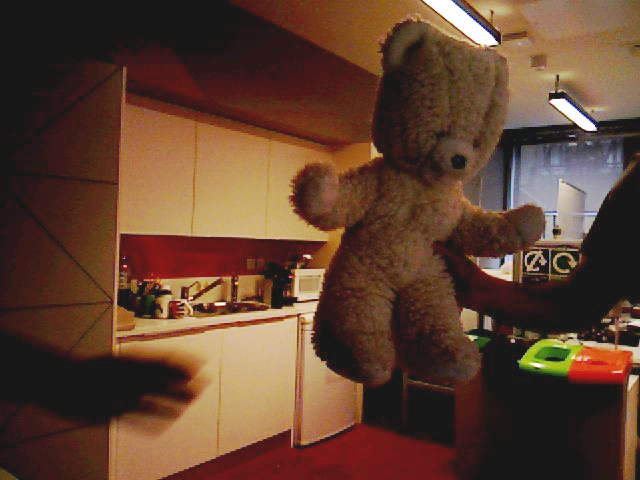} &
\includegraphics[width=\imwidth\linewidth]{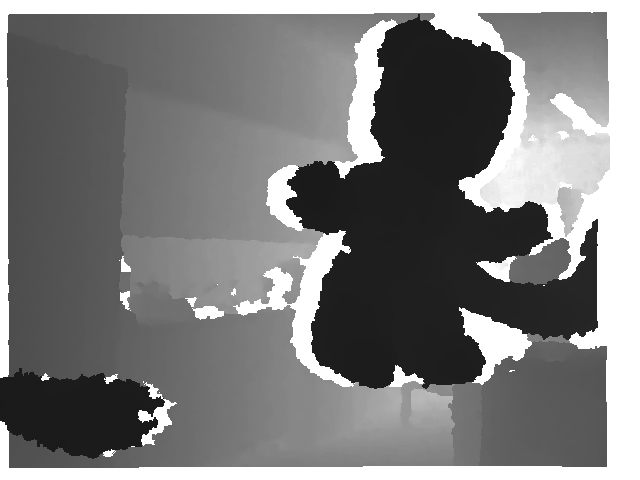} &
\includegraphics[width=\imwidth\linewidth]{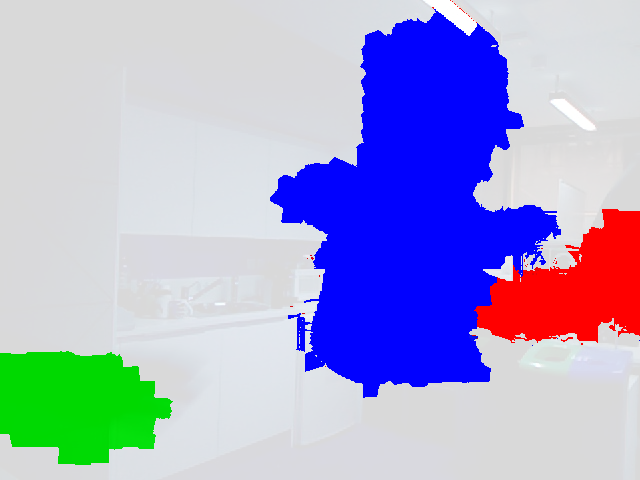} &
\includegraphics[width=\imwidth\linewidth]{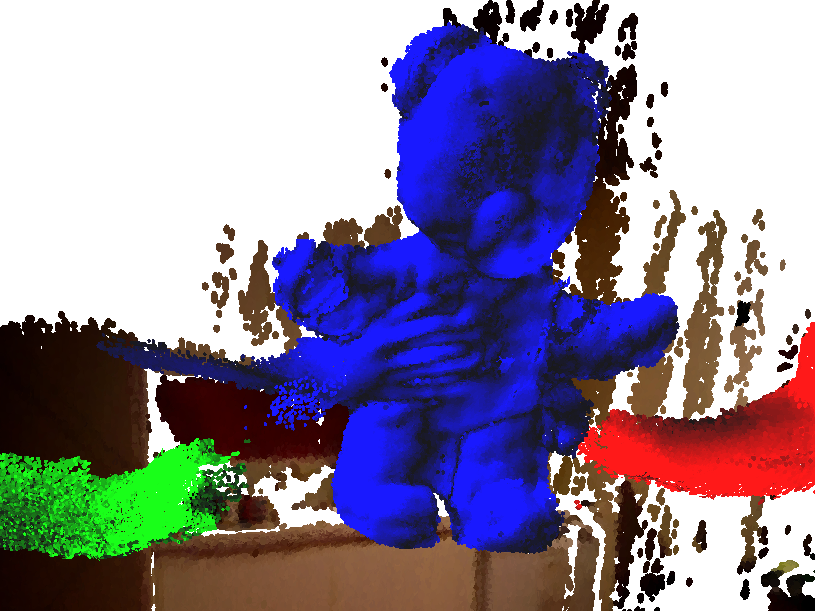} \\

\end{tabular}
\caption{In this \emph{handover} sequence a toy teddy bear is handed from
one person to another. Co-Fusion can correctly
segment and model four bodies: The background, the teddy-bear and two
arms. At the start, the left arm and teddy are represented by the same
model, since they move together. When the handover occurs, however,
the arm becomes separated from the teddy and all four objects are
tracked independently.}
\label{fig:handover}
\end{figure*}

Once detected and segmented, objects are added to the list
of \emph{active} models and are subsequently tracked and their 3D
shape model updated by fusing only the data labeled as belonging to
that object. The tracking and fusion threads for each object are based
on recent surfel-based approaches~\cite{surfel-fusion,elastic-fusion}.
The main contribution of this paper is a system that
would allow a robot not only to reconstruct its surrounding
environment but also to acquire the detailed 3D geometry of unknown
objects that move in the scene. Moreover, our system would equip a
robot with the capability to discover new objects in the scene and
learn accurate 3D models of them through active motion. We demonstrate
Co-Fusion on different scenarios -- placing different
previously unseen objects on a table and learning their geometry (see
Figure~\ref{fig:placed-items}), handing over an object from one person
to another (see Figure~\ref{fig:handover}),
%capturing articulated motion (see figure~\ref{fig:articulated}),
hand-held 3D capture of a
moving object with a moving camera (see
Figure~\ref{fig:teddy-reconstruction}) and on a car driving scenario
(see Figure~\ref{fig:semantic-street}). We also demonstrate
quantitatively the robustness of the tracking and the reconstruction
on some synthetic and ground truth sequences of dynamic scenes.

\section{RELATED WORK}

The arrival of the Microsoft Kinect device and the sudden availability
of inexpensive depth sensors to consumers, triggered a flurry of
research aimed at real-time 3D scanning. Systems such as
KinectFusion~\cite{kinect-fusion} first made it possible to map the 3D
geometry of arbitrary indoor scenes accurately and in real time, by
fusing the images acquired by the depth camera simply by moving the
sensor around the environment. Access to accurate and \emph{dense} 3D
geometry in \emph{real time} opens up applications to rapid scanning
or prototyping, augmented/virtual reality and mobile robotics that
were previously not possible with offline or sparse
techniques. Successors to KinectFusion have quickly addressed some of
its shortcomings. While some have focused on extending its
capabilities to handle very large
scenes~\cite{henry-fox,whelan-iros13,niessner-tog-13,kintinuous} or to
include loop closure~\cite{elastic-fusion} others have robustified the
tracking~\cite{kintinuous} or improved memory and scale efficiency by
using point-based instead of volumetric
representations~\cite{surfel-fusion} that lead to increased 3D
reconstruction quality~\cite{anisotropic}. Achieving higher level
semantic scene descriptions by using a dense planar
representation~\cite{planar-slam} or real-time 3D object
recognition~\cite{slam++} further improved tracking performance while
opening the door to virtual or even real  interaction with the scene. More recent approaches such as~\cite{tateno2016icra,li2016iros} incorporate semantic segmentation and even recognition within a SLAM system in real time. While they show impressive performance, they are still limited to static scenes.

The core underlying assumption behind many traditional SLAM and dense
reconstruction systems is that the scene is largely static. How can
these dense systems be extended to track and reconstruct more than one
model without compromising real time performance? The SLAMMOT
project~\cite{slammot} represented an important step towards extending
the SLAM framework to dynamic environments by incorporating the
detection and tracking of moving objects into the SLAM operation. It
was mostly demonstrated on driving scenarios and limited to sparse
reconstructions.  It is only very recently that the problem of
reconstruction of \emph{dense} dynamic scenes in real time has been
addressed. Most of the work has been devoted to capturing non-rigid
geometry in real time with RGB-D sensors. The assumption here is that
the camera is observing a single object that deforms freely over
time. DynamicFusion~\cite{dynamic-fusion} is a prime example of a
monocular real time system that can fuse together scans of deformable
objects captured from depth sensors without the need for any
pre-trained model or shape template. With the use of a sophisticated
multi-camera rig of RGB-D sensors 4DFusion~\cite{fusion4d} can capture
live deformable shapes with an exceptional level of detail and can
deal with large deformations and changes in topology. On the other
hand template based techniques can also obtain high levels of realism
but are limited by their need to add a preliminary step to capture the
template~\cite{templatefusion} or are dedicated to tracking specific
objects by their use of hand-crafted or pre-trained
models~\cite{hand-msr}.  These include general articulated tracking
methods that either require a geometric template of the object in a
rest pose~\cite{gall}, or prior knowledge of the skeletal
structure~\cite{dart}.

In contrast, capturing the full geometry of dynamic scenes that might
contain more than one moving object has received more limited
attention. Ren \emph{et al.}~\cite{Ren:etal:iccv13} propose a method
to track and reconstruct 3D objects simultaneously by refining an
initial simple shape primitive.  However, in contrast to our approach,
it can only track one moving object and requires a manual
initialization.~\cite{Martin-Martin:etal:icra16} propose a combined
approach for estimating pose, shape, and the kinematic structure of
articulated objects based on motion segmentation. While it is also
based on joint tracking and segmentation, the focus is on discovering
the articulated structure, only foreground objects are reconstructed
and its performance is not real time. St{\"{u}}ckler and
Behnke~\cite{stuckler} propose a dense rigid-body motion segmentation
algorithm for RGB-D sequences. They only segment the RGB-D images and
estimate the motion but do not simultaneously reconstruct the
objects. Finally~\cite{Herbst:etal:icra14} build a model of the
environment and consider as new objects parts of the scene that become
inconsistent with this model using change detection. However, this
approach requires a human in the loop to acquire known-correct
segmentation and does not provide real time operation.

Several recent RGB-only methods have also addressed the problem of
monocular 3D reconstruction of dynamic scenes.  Works such
as~\cite{video-popup,fragkiadaki,dense-multibody} are similar in
spirit to our simultaneous segmentation, tracking and reconstruction
approach. Russell \emph{et al.}~\cite{video-popup} perform multiple
model fitting to decompose a scene into piecewise rigid parts that are
then grouped to form distinct objects. The strength of their approach
is the flexibility to deal with a mixture of non-rigid, articulated or
rigid objects. Fragkiadaki \emph{et al.}~\cite{fragkiadaki} follow a
pipeline approach that first performs clustering of long term tracks
into different objects followed by non-rigid reconstruction. However,
both of these approaches act on sparse tracks and are batch methods
that require all the frames to have been captured in advance.
Our method also shares commonality with the dense RGB multi-body
reconstruction approach of~\cite{dense-multibody}, who also perform
simultaneous segmentation, tracking and 3D reconstruction of multiple
rigid models, with the notable difference that our approach is online
and real time while theirs is batch and takes several seconds per
frame.

%% %%%%%%%%%%%%%%%%%%%%%%%%%%%%%%%%%%%%%%%%%%%%%%%%%%%%%%%%%%%%%%%%%%%%%%%%%%%%%%%%
\begin{figure}[tbp]
	\centering
	\includegraphics[width=.8\linewidth]{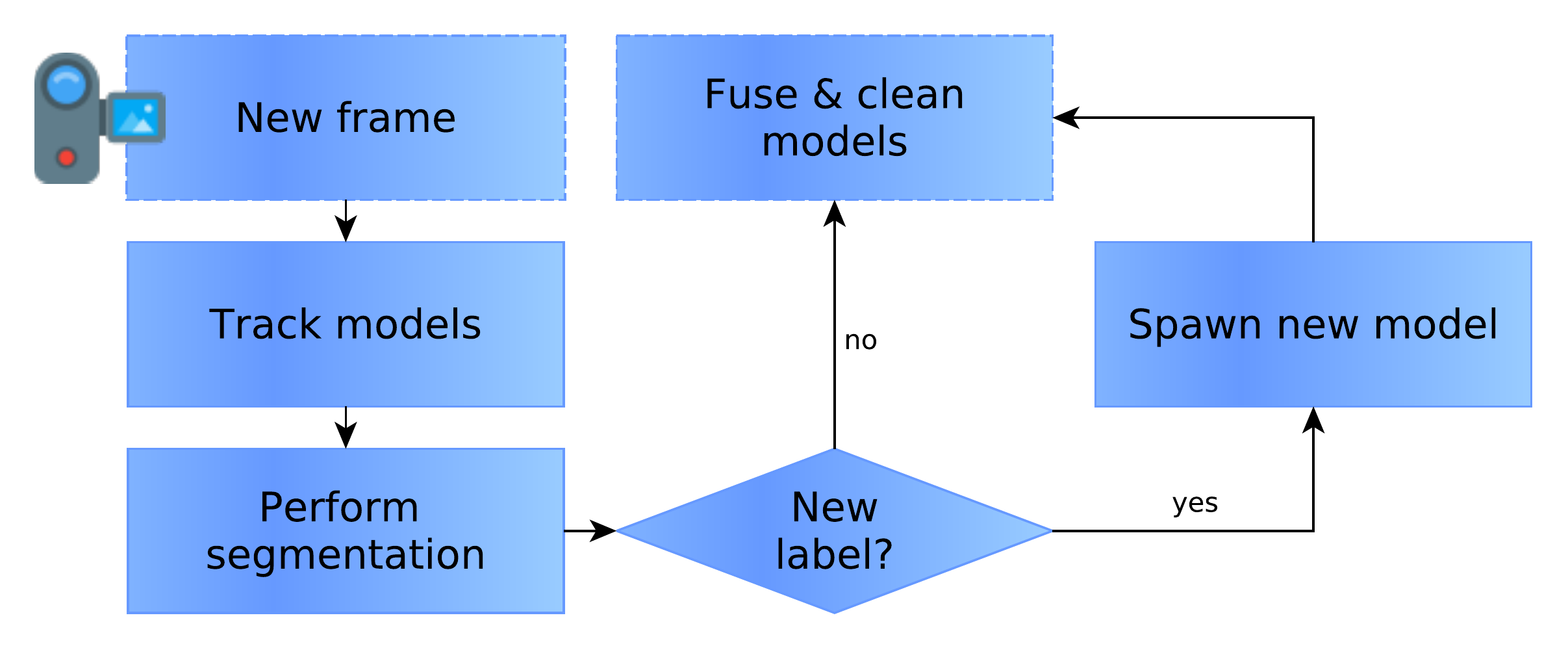}
	\caption{Overview of our method showing the data-flow starting
	from a new RGBD-frame. A detailed description can be found in Section~\ref{sec:overview} }
	\label{fig:pipeline_overview}
\end{figure}

\section{OVERVIEW OF OUR METHOD}\label{sec:overview}

Co-Fusion is a live RGB-D SLAM system that processes each new frame in
real time. As well as maintaining a global model of the detailed
geometry of the background our system stores models for each object
segmented in the scene and is capable of tracking their motions
independently. Each model is stored simply as a set of 3D points. Our
system maintains two sets of object models: while \emph{active} models
are objects that are currently visible in the live
frame, \emph{inactive} models are objects that were once visible,
therefore their geometry is known, but are currently out of view.

Figure~\ref{fig:pipeline_overview} illustrates the frame-to-frame
operation of our system. At the start of live capture, the scene is
initialized to contain a single \emph{active} model -- the
background. Once the fused 3D model of the background and the camera
pose are stable after a few frames our system follows the pipeline
approach described below. For each new frame acquired by the camera
the following steps are performed:

\noindent {\bf Tracking} First, we track
the 6DOF rigid pose of each \emph{active} model in the current
frame. This is achieved by minimizing an objective function
independently for each model that combines a geometric error based on
dense iterative closest point (ICP) alignment and a photometric cost
based on the difference in color between points in the current live
frame and the stored 3D model.

\noindent {\bf Segmentation} In this step we segment the
current live frame associating each of its pixels with one of the
active models/objects. Our system can perform segmentation based on
two different cues: \emph{(i)} motion and \emph{(ii)} semantic
labels. We now describe each of these two grouping strategies.

\noindent \emph{(i) Motion segmentation}
We formulate motion segmentation as a labeling problem using a
fully connected Conditional Random Field and optimize it in real time
on the CPU with the efficient approach of~\cite{crf}. The unary
potentials encode the geometric ICP cost incurred when associating a
pixel with a rigid motion model.
The optimization is followed by the extraction of connected components
in the segmented image. If the connected region occupied by outliers
has sufficient support
an object is assumed to have entered the scene and a new model is
spawned and added to the list.

\noindent \emph{(ii) Multi-class image segmentation} As an alternative
to motion segmentation our system can segment object instances at the
pixel level given a class label using an efficient state of the art
approach~\cite{sharp-mask} based on deep learning. This allows us to
segment objects based on semantic cues. For instance, in an autonomous
driving application our system could segment not just moving but
also stationary cars.

\noindent {\bf Fusion} Using the newly estimated 6-DOF pose, the dense
3D geometry of each \emph{active} model is updated by fusing the
points labeled as belonging to that model. We used a surfel-based
fusion approach related  to the methods
of \cite{surfel-fusion} and \cite{elastic-fusion}.

While the tracking and fusion steps of our pipeline run on the GPU,
the segmentation step runs on the CPU. The result is an
RGB-D SLAM system that can maintain an up-to-date 3D map of the static
background as well as detailed 3D models for up to 5 different objects at 12
frames per second.

\section{NOTATION AND PRELIMINARIES}

We use $\Omega$ to refer to the 2D image domain that contains all the
valid image coordinates. These are denoted as ${\bf u}= (u_x,u_y)^T \in
\Omega$ and their homogeneous coordinates as $\dot{{\bf u}}=({\bf u}^T,1)^T$. An RGB-D frame contains both a depth image $\mathcal D$ of
depth pixels $d({\bf u}): \Omega \rightarrow \mathbb{R}$ and an RGB
image $\mathcal C$ of color pixels $c({\bf
u}): \Omega \rightarrow \mathbb{N}^3$.  The greyscale intensity value
of pixel ${\bf u}$ given color  $c({\bf
u})=[c_r,c_g,c_b]$ in image $\mathcal{C}$ is given by ${\bf I}({\bf u}) =
{{(c_r+c_g+c_b)}\over{3}} \in \mathbb{R}$.
The perspective projection of a 3D point ${\bf p} = (x, y, z)^T$ is specified as
${\bf u} = \pi({\bf K}{\bf p})$ where $\pi
: \mathbb{R}^3 \rightarrow \mathbb{R}^2$ $\pi({\bf
p})=(x/z,y/z)^T$. The back-projection of a point ${\bf u} \in
\Omega$ given its depth $d({\bf u})$ can be expressed as $\pi^{-1}({\bf u},\mathcal D)
= {\bf K}^{-1} \dot{{\bf u}} d({\bf u}) \in  \mathbb{R}^3$.

Similarly to~\cite{surfel-fusion} and~\cite{elastic-fusion}, we use a
surfel-based map representation. For each \emph{active}
and \emph{inactive} model a list of unordered surfels $\mathcal{M}_m$
is maintained, where each surfel ${\mathcal{M}_m^s \in
(\mathbf{p}\in\mathbb{R}^3, \mathbf{n}\in\mathbb{R}^3, \mathbf{c}\in\mathbb{N}^3, \mathbf{w}\in\mathbb{R}, \mathbf{r}\in\mathbb{R}, \mathbf{t}\in \mathbb{R}^2)}$
is a tuple of position, normal, color, weight, radius and two
timestamps.

Given that we are modeling dynamic scenes where not just the camera
but other objects might move, we use $\mathcal T_t=\{{\bf
T_{tm}}(\cdot)\}$ to describe the the set of $M_t$ rigid
transformations that encode the pose of each \emph{active} model
$\mathcal{M}_m$ at time instant $t$ with respect to the global
reference frame. In other words, ${\bf T_{tm}}$ is the rigid transform
${\bf T_{tm}}({\bf p}_m) = \bR_{tm}{\bf p}_m +\bt_{tm}$, that aligns a
3D point ${\bf p}_m$ lying on model $m$ expressed in the global
reference frame, to its current position at time $t$.
$\bR_{tm} \in \mathbb{SO}_3$ and $\bt_{tm} \in \mathbb{R}^3$ are
respectively the rotation matrix and translation vector. We reserve
the notation ${\bf T_{tb}}$ to refer specifically to the rigid
transforms associated with the background model.

%%%%%%%%%%%%%%%%%%%%%%%%%%%%%%%%%%%%%%%%%%%%%%%%%%%%%%%%%%%%%%%%%%%%%%%%%%%%%%%%
\section{TRACKING ACTIVE MODELS}

\begin{figure}
\center
\def \imwidth {.22}
\settoheight{\tempdima}{\includegraphics[width=\imwidth\linewidth]{handover/before-color}}
\begin{tabular}{@{}c@{ }c@{ }c@{ }c@{ }c@{}}

&background&teddy&arm 1&arm 2\\
\rowname{Before} &
\includegraphics[width=\imwidth\linewidth]{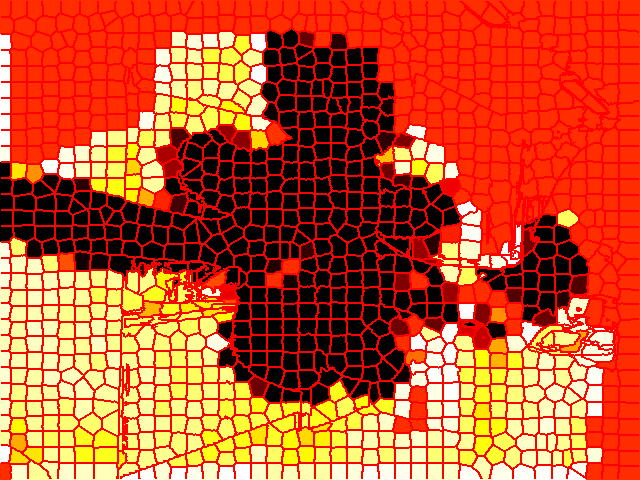} &
\includegraphics[width=\imwidth\linewidth]{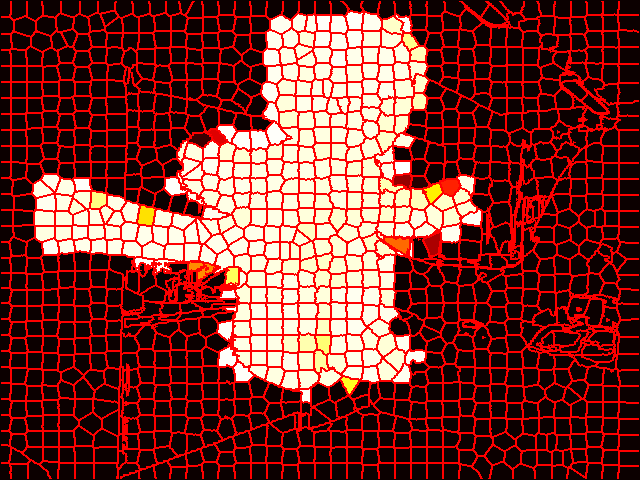} &
\includegraphics[width=\imwidth\linewidth]{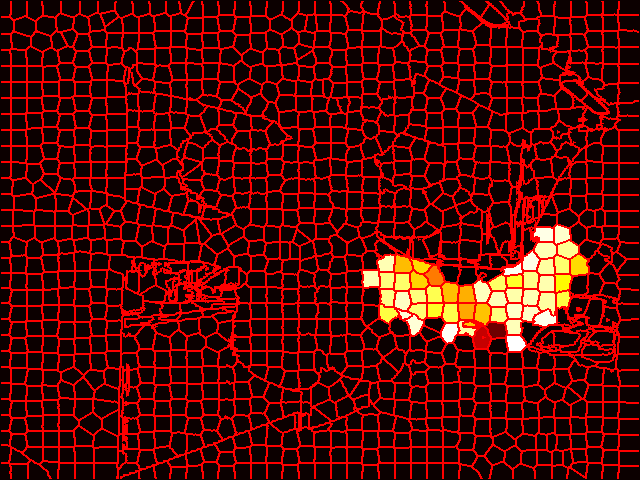} &
\includegraphics[width=\imwidth\linewidth]{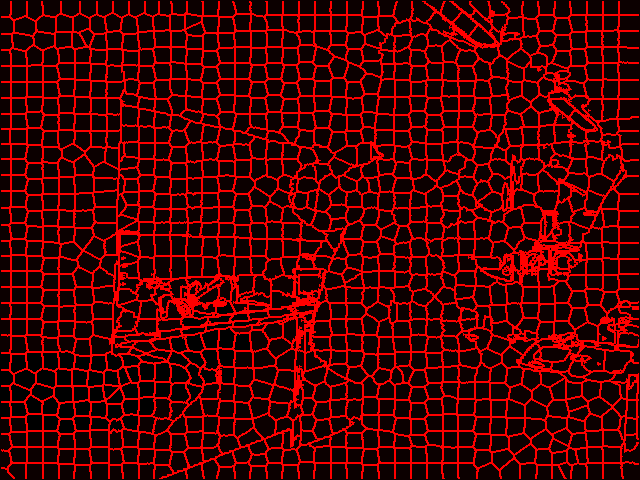} \\

\rowname{After} &
\includegraphics[width=\imwidth\linewidth]{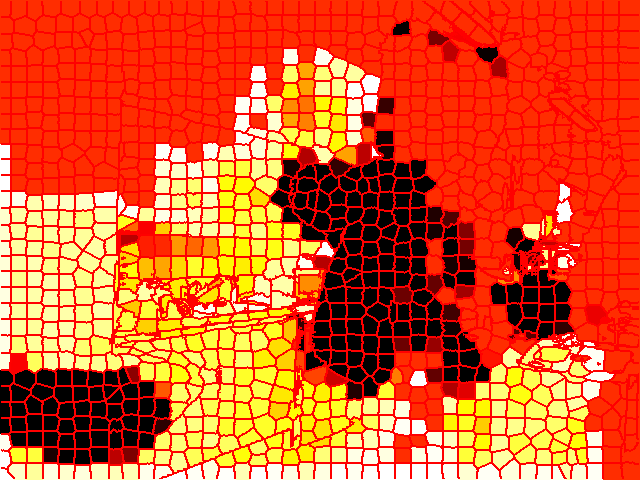} &
\includegraphics[width=\imwidth\linewidth]{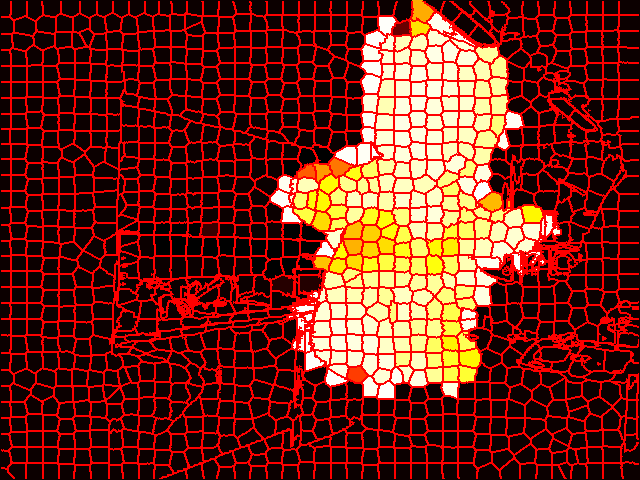} &
\includegraphics[width=\imwidth\linewidth]{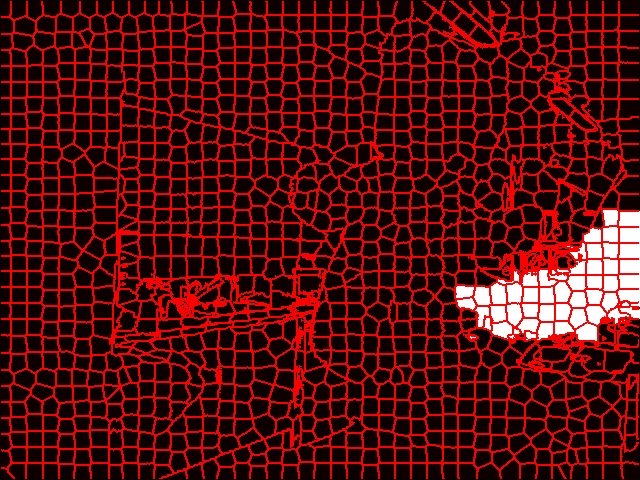} &
\includegraphics[width=\imwidth\linewidth]{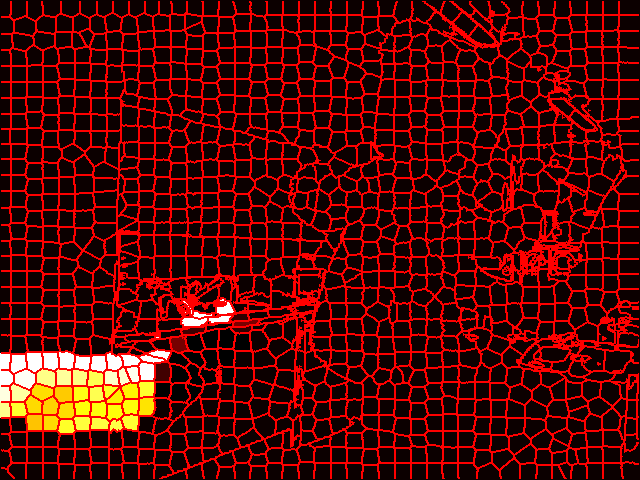} \\

\end{tabular}
\caption{Heat-map visualization of the \emph{unary potentials} for each of
the four model labels in the \emph{handover} scene (see Figure~\ref{fig:handover}). Brighter values correspond to a higher probability of each label
being assigned to a super-pixel.}
\end{figure}

For each input frame at time $t$ and for each \emph{active} model
 $\mathcal{M}_m$ we track its global pose ${\bf T_{tm}}$ by
 registering the current live depth map with the predicted depth map
 in the previous frame, obtained by projecting the stored 3D model
 using the estimated pose for $t-1$. We track each \emph{active} model
 independently by running the optimization described below selecting
 only the 3D map points that are labeled as belonging to that
 specific model.

\subsection{Energy}

For each \emph{active} model $\mathcal{M}_m$, we minimize a cost
function that combines a geometric term based on point-to-plane ICP
alignment and a photometric color term that minimizes differences in
brightness between the predicted color image resulting from projecting
the stored 3D model in the previous frame and the current live color
frame.

\begin{equation}
E^m_{track} = \min_{{\bf T_{m}}}\,\, (E^m_{icp} + \lambda E^m_{rgb})
\end{equation}

 This cost function is closely related to the tracking threads of
 other RGB-D based SLAM
 systems~\cite{elastic-fusion,surfel-fusion}. However, the most
 notable difference is that while~\cite{elastic-fusion,surfel-fusion}
 assume that the scene is static and only track a single model,
 Co-Fusion can track various models while maintaining real-time
 performance.

\subsection{Geometry Term}

For each \emph{active} model $m$ in the current frame $t$ we seek to
minimize the cost of the \emph{point-to-plane ICP} registration error
between \emph{(i)} the 3D back-projected vertices of  the current live
depth map and \emph{(ii)} the predicted depth map of model $m$ from
the previous frame $t-1$:
\begin{equation}\label{eqn:icp_error}
E^m_{icp} = \sum_i(({\bf v^i} - {\bf T_m}{\bf v_t^i})\cdot{\bf n^i})^2
\end{equation}
where ${\bf v_t^i}$ is the back-projection of the $i$-th vertex in the
current depth-map $\mathcal{D}_t$; and ${\bf v^i}$ and ${\bf n^i}$ are
respectively the back-projection of the $i$-th vertex of the predicted
depth-map of model $m$ from the previous frame $t-1$ and its normal.
${\bf T_m}$ describes the transformation that aligns  model $m$ in
the previous frame $t-1$ with the current frame $t$.

\subsection{Photometric Color Term}

Given \emph{(i)} the current depth image; \emph{(ii)} the current
estimate of the 3D geometry of each \emph{active} model;
and \emph{(iii)} the estimated rigid motion parameters that align each
model with respect to the previous frame $t-1$, it is possible to
synthesize projections of the scene onto a virtual camera aligned with
the previous frame.

The tracking problem then becomes one of photometric image
registration where we minimize the brightness constancy between the live
frame and the synthesized view of the 3D models in frame $t-1$.

The cost takes the form

\begin{equation}\label{eqn:photometric_error}
E^m_{rgb}= \sum_{{\bf u} \in \Omega_m} ({\bf I_t}({\bf u})-{\bf I_{t-1}}(\pi({\bf K}{\bf T_m}\pi^{-1}({\bf u},\mathcal{D}_t))^2
\end{equation}
where ${\bf T_m}$ is the rigid transformation that aligns
\emph{active} model $\mathcal{M}_m$ between the previous frame $t-1$ and the
current frame and ${\bf I_{t-1}(\cdot)}$ is a function that provides the
color attached to a vertex on the model in the previous frame $t-1$.

For reasons of robustness and efficiency this optimization is embedded
in a coarse-to-fine approach using a 4-layer spatial pyramid. Our GPU
implementation builds on the open source code release
of~\cite{elastic-fusion}.

%%%%%%%%%%%%%%%%%%%%%%%%%%%%%%%%%%%%%%%%%%%%%%%%%%%%%%%%%%%%%%%%%%%%%%%%%%%%%%%%
\section{MOTION SEGMENTATION}

Following the tracking step we have new estimates for the $M_t$ rigid
transformations $\{{\bf T_{tm}}\}$ that describe the absolute pose of
each \emph{active} model with respect to the global reference frame at
time $t$.

We now formulate the motion segmentation problem for a new input frame
$t$ as a labeling problem, where the labels are the $M_t$ rigid
transformations $\{{\bf T_{tm}}\}$. We seek a labeling ${\bf x}({\bf
u}):\Omega \rightarrow {\mathcal L}_t$ that assigns a label $\ell \in
{\mathcal L}_t=\{1,\dots,\vert M_t \vert + 1\}$ to each point ${\bf u}$ in the current
frame associating it with the motion of one of the $M_t$
currently \emph{active} rigid models or an outlier label $\ell_{\vert M_t\vert+1}$. Note that the number
of \emph{active} models (labels) $M_t$ will vary per frame as new objects
may appear or disappear in the scene.

In practice, to allow the motion segmentation to run in real time on
the CPU, we first over segment the current frame into SLIC
super-pixels~\cite{slic} using the fast implementation of~\cite{gslicr}
and apply the labeling algorithm at the super-pixel level. The
position, color and depth of each super-pixel is estimated by
averaging those of the pixels inside it.

We follow the energy minimization approach of~\cite{crf} that
optimizes the following cost function with respect to the labeling
${\bf x_t} \in {\mathcal L}^S$
\begin{equation}
E({\bf x_t}) = \sum_i \psi_u(x_i) + \sum_{i<j} \psi_p(x_i,x_j)
\end{equation}

where $i$ and $j$ are indices over the image super-pixels ranging from
 $1$ to $S$ (the total number of super-pixels).

\noindent {\bf The unary potentials $\psi_u(x_i)$} denote the cost
 associated with a label assignment $x_i$ for super-pixel ${\bf
 s}_i$. Given that we are solving a motion segmentation problem, the
 unary potentials are the estimated {\sc icp} alignment costs incurred
 when applying the rigid transformation associated with each label
 %(\emph{active} model) $\mathcal{M}_m$
to the back-projection of the center
 of each super-pixel $s_i$ as defined in \eqref{eqn:icp_error}. Note
 that this is a purely geometric cost.
If computing the cost $\psi_u(x_i)$ fails due to lack of geometry
projecting to $s_i$, we assign a fixed cost corresponding to a
misalignment of 1\% of the depth-range of the current frame. This
 prevents labels from growing outside of the object
bounds. For each super-pixel, the unary cost associated with the
outlier label $\ell_{\vert M_t\vert+1}$
%= \phi - min(\ell_0,\dots,\ell_{\vert M_t\vert})$
is determined by the cost of the best fitting label and as a result
receives low values only if none of the rigid models can explain the motion
of the super-pixel.% \hl{ups, fix inconsistent notation here}
%std::max(unaryThresholdNew - unaryWeightError * lowestError, 0.01f);

%
\noindent {\bf The pairwise potentials $\psi_p(x_i,x_j)$} can be expressed as
\begin{equation}
 \psi_p(x_i,x_j) = \mu(x_i,x_j) \sum_{m=1}^K \omega_m k_m(f_i,f_j).
% \psi_p(x_i,x_j) = \mu(x_i,x_j) (\omega_1 k_1(f_i,f_j)+\omega_2 k_2(f_i,f_j))
\end{equation}
where $\mu(x_i,x_j)$ encapsulates the classic Potts model that
penalizes nearby pixels taking different labels, and $k_m(f_i,f_j)$
are contrast-sensitive potentials that measure the similarity between
the appearance of pixels. This results in a cost that
encourages super-pixels $i$ and $j$ to take the same label if the
distance between their feature vectors $f_i$ and $f_j$ is small. In
practice we characterize each super-pixel $i$ with the 6D feature
vector $f_i$ that encodes its 2D location, RGB color and depth
value. We set $k_m$ to be Gaussian kernels

$k_m(f_i,f_j) = \exp(-{{1}\over{2}}(f_i - f_j)^T \Lambda_m (f_i -
f_j))$  with $\Lambda_m $ the inverse covariance matrix \footnote{
In practice we set $K=2$ and the inverse covariance matrices to
$\Lambda_1=$diag$(1/\theta_{\alpha}^2,1/\theta_{\alpha}^2,1/\theta_{\beta}^2,1/\theta_{\beta}^2,1/\theta_{\beta}^2,1/\theta_{\gamma}^2)$
and $\Lambda_2=$diag$(1/\theta_{\delta}^2,1/\theta_{\delta}^2,0,0,0,0)$}.

We use the efficient inference method of~\cite{crf} to optimize the
labeling, which can be computed in real time on the {\sc CPU}. The
output of this optimization is a soft assignment of labels to each
super-pixel $i$. To convert this into a hard assignment we simply take
the maximum of all the label assignments and associate each super-pixel
with the motion of a single \emph{active} model.

{\bf Post-processing.} Following the segmentation we
perform a series of post-processing steps to obtain more robust
results. First we perform connected components for all the labels and
we merge models that have similar rigid transformations.
Secondly we ensure that disconnected regions are modeled separately
by suppressing all except the largest component with the same
label. In a similar way, components whose size falls below a threshold
$\tau$ are removed.

%%%%%%%%%%%%%%%%%%%%%%%%%%%%%%%%%%%%%%%%%%%%%%%%%%%%%%%%%%%%%%%%%%%%%%%%%%%%%%%%
\subsection{Addition of New Models}

\begin{figure*}
  \def \imwidth {.24}
  \centering
  \subfloat[Color image]{\includegraphics[width=\imwidth\linewidth]{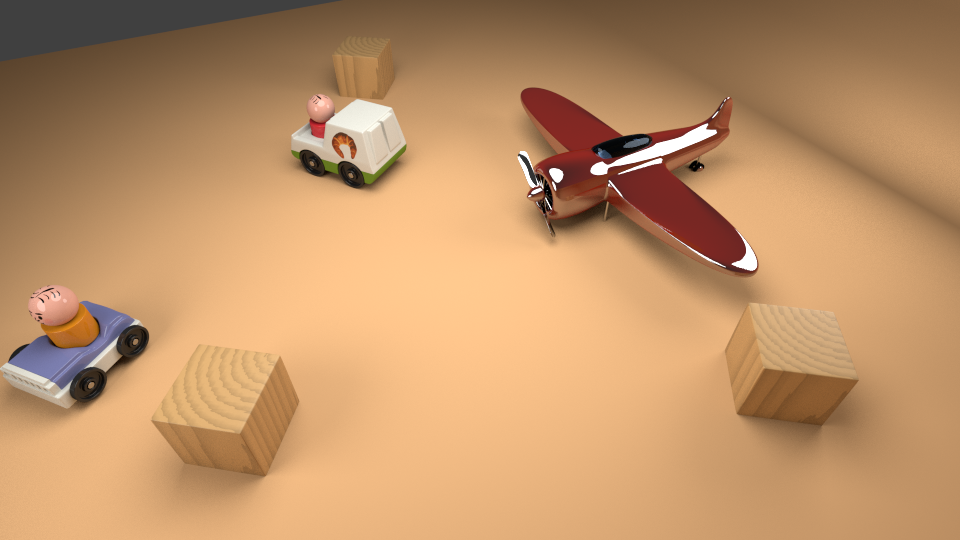}}~
  \subfloat[Motion segmentation]{\includegraphics[width=\imwidth\linewidth]{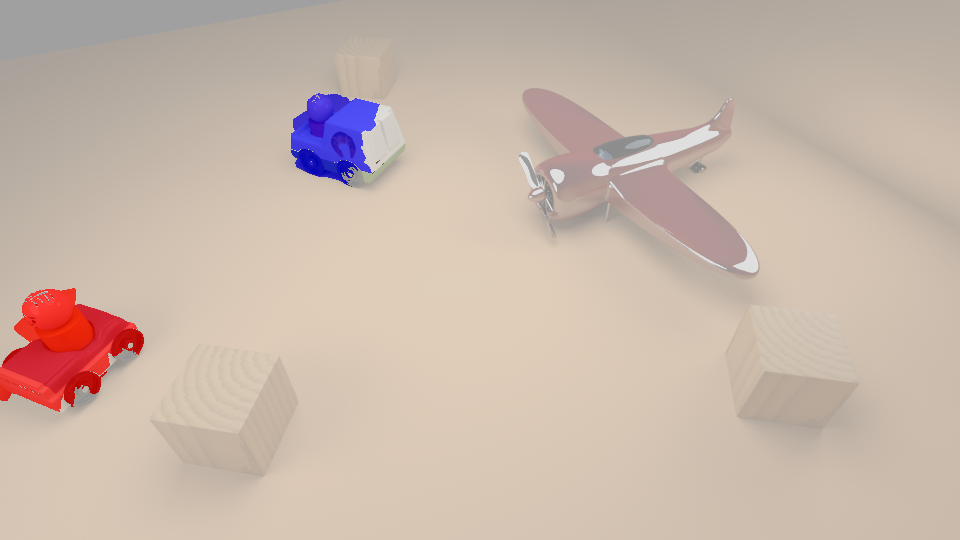}}~
  \subfloat[Colored 3D models]{\includegraphics[width=\imwidth\linewidth]{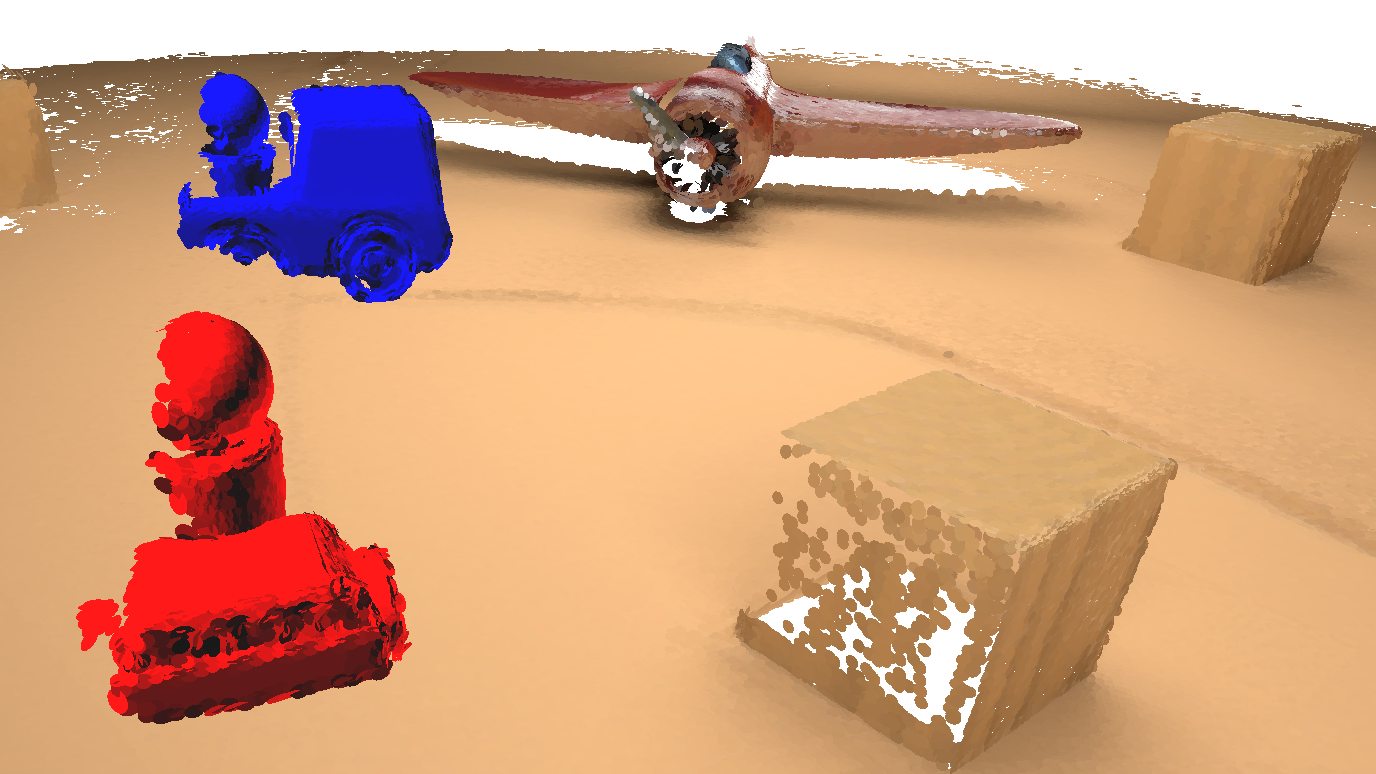}}~
  \subfloat[Reconstruction result]{\includegraphics[width=\imwidth\linewidth]{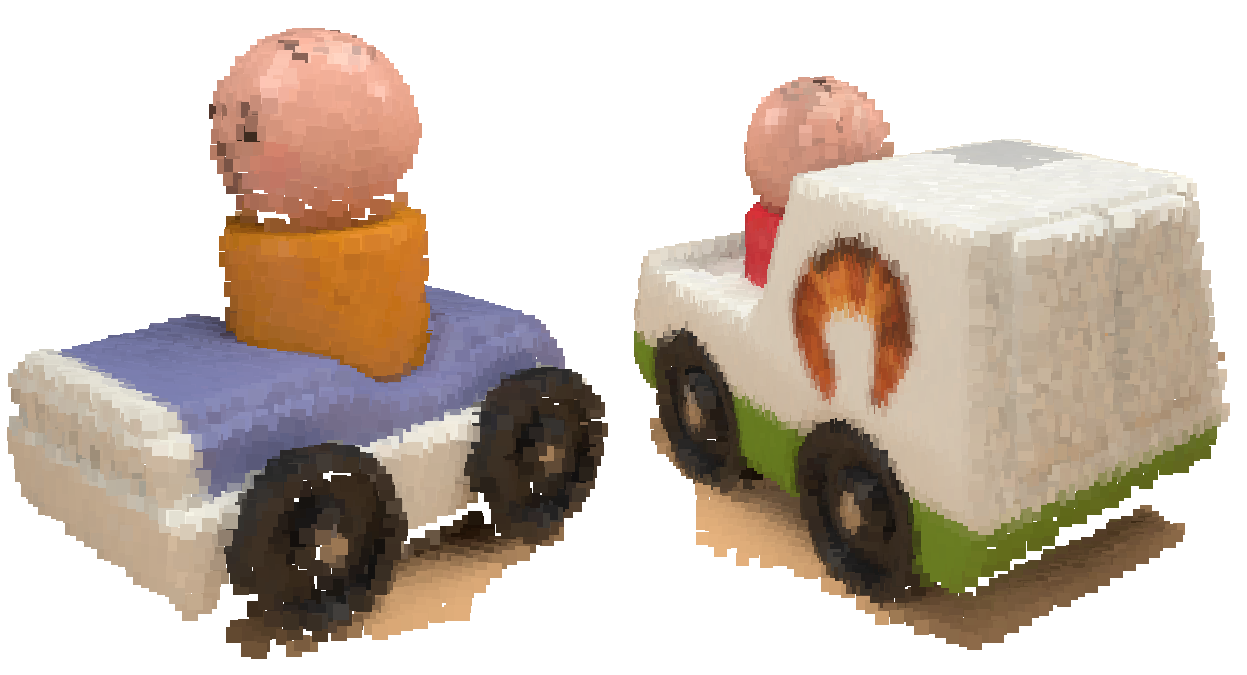}}
  \setlength{\belowcaptionskip}{-17pt}
  \caption{Visualization of the stages of Co-Fusion based on our synthetic \emph{ToyCar3} sequence.}
  \label{fig:teddy-back-removal}
\end{figure*}

If the connected region occupied by outliers is larger than $3\%$ of
the total number of pixels, an object is assumed to have entered the
scene and a new label/object is spawned. If part of the geometry of
this new object was already in the map (for instance, if an object
started moving after having been part of the background map for a
while) we attempt to remove the duplicate reconstruction. In practice
we found that a good strategy is to remove areas with a high ICP error
from the background. This is illustrated in
Figure~\ref{fig:teddy-back-removal}.

On the other hand, if a label disappeared and does not reappear within
a certain number of frames, it is assumed that the respective model
left the scene. In this case the model will be added to
the \emph{inactive} list, if it contains enough surfels with a high
confidence and is deleted otherwise.

%%%%%%%%%%%%%%%%%%%%%%%%%%%%%%%%%%%%%%%%%%%%%%%%%%%%%%%%%%%%%%%%%%%%%%%%%%%%%%%%
\section{OBJECT INSTANCE SEGMENTATION}

\begin{figure*}
\centering
\def \imwidth {.99}
%\subfloat[Semantic labels]{\includegraphics[width=\imwidth\linewidth]{semantic/street1-deep-fullscaled}\label{fig:semantic-street}} \\
\subfloat[Semantic labels]{\includegraphics[width=.4\linewidth]{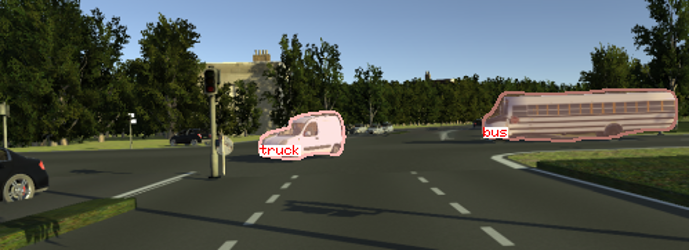}\label{fig:semantic-street}}
%\subfloat[Labels in 3D (front view)]{\includegraphics[width=\imwidth\linewidth]{semantic/street1-incar3}} \\
\subfloat[Labels in 3D (front view)]{\includegraphics[width=.3\linewidth]{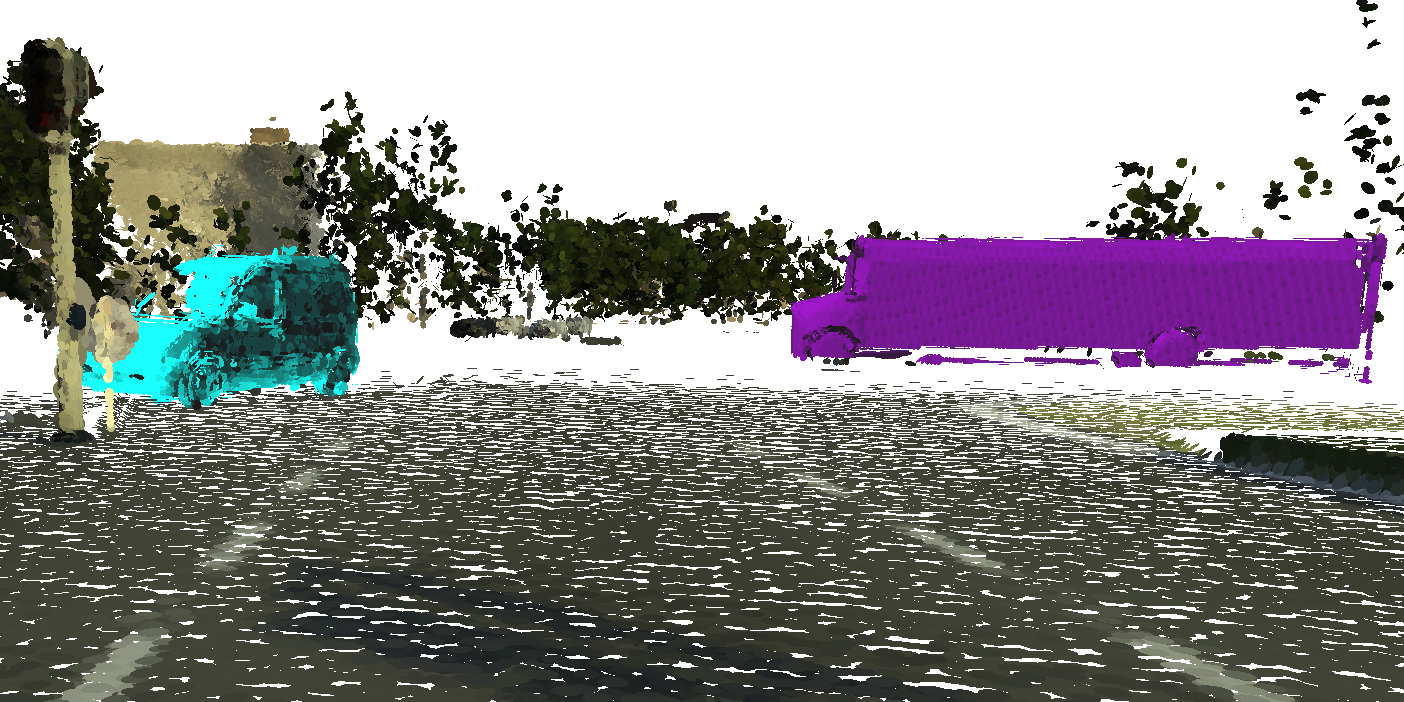}}
%\subfloat[3D reconstruction (top view)]{\includegraphics[width=\imwidth\li%newidth]{semantic/street1-top-color2-1}} \enspace
%\subfloat[Labels in 3D (top view)]{\includegraphics[width=\imwidth\linewidth]{semantic/street1-top-segmentation2-1}}
\subfloat[Labels in 3D (top view)]{\includegraphics[width=.3\linewidth]{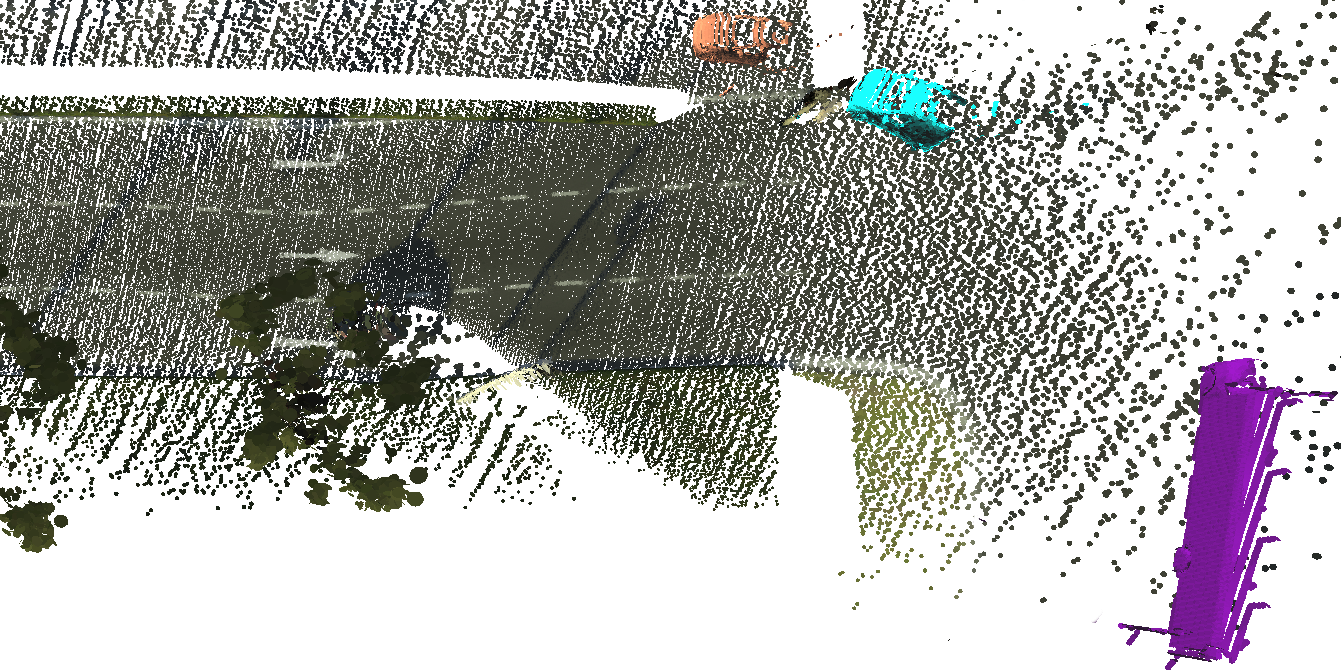}}
\setlength{\belowcaptionskip}{-5pt}
\caption{Results based on the semantic labeling. Here we show a scene from the virtual KITTI dataset \cite{virtual-kitti},
which would be difficult for our motion based segmentation. While \ref{fig:semantic-street} shows semantic labels generated by a CNN,
the remaining images show the reconstruction and highlight the object labels.}
\end{figure*}

In this section we investigate the use of semantic cues to segment
 objects in the scene which allows to deal both with moving and static
 objects.  We use the top performing state of the art method for
 object instance segmentation~\cite{sharp-mask} to segment objects of
 interest. \emph{SharpMask}~\cite{sharp-mask} is an augmented
 feed-forward network able to predict object proposals and object
 masks simultaneously. The architecture has 3 elements: A pre-trained
 network for feature map extraction, a segmentation branch and a
 branch that scores the `objectness' of an image patch.
The results of \emph{SharpMask} (an example segmentation can be seen
in Figure~\ref{fig:semantic-street}) can be given directly to
Co-Fusion after temporal consistency is imposed between consecutive
frames. The segmentation can be run on a limited  set of
labels to segment only objects of a
chosen class, for instance all the tools lying on a table.
We used the publicly available models pre-trained on
the COCO dataset~\cite{coco-dataset}.

%%%%%%%%%%%%%%%%%%%%%%%%%%%%%%%%%%%%%%%%%%%%%%%%%%%%%%%%%%%%%%%%%%%%%%%%%%%%%%%%
\section{FUSION}\label{sec:fusion}

During the tracking stage, \emph{active} models
$\mathcal{M}_m$ are projected to the camera view using
splat rendering in order to align individual model poses. In the
subsequent fusion stage the surfel maps are updated by merging the
newly available RGB-D frame into the existing models. After
projectively associating image coordinates ${\bf u}$ with
corresponding surfels in the model $\mathcal{M}_m$, an update scheme
similar to~\cite{surfel-fusion} is used.

%%%%%%%%%%%%%%%%%%%%%%%%%%%%%%%%%%%%%%%%%%%%%%%%%%%%%%%%%%%%%%%%%%%%%%%%%%%%%%%%

\section{Evaluation}

We carried out a
quantitative evaluation both on synthetic and  real
sequences with ground truth data.  Appropriate {\bf synthetic
sequences} with Kinect-like noise~\cite{handa:etal:ICRA2014} were
specifically created for this work (\emph{ToyCar3} and \emph{Room4})
and have been made publicly available, along with evaluation
tools. For the {\bf ground truth experiments on real data} we attached
markers to a set of objects, as shown in Figure~\ref{fig:esone}, and
accurately reconstructed them using a NextEngine 3D-scanner.  The scenes were recorded with a motion-capture system (OptiTrack) to obtain ground-truth data for the trajectories. An Asus
Xtion was used to acquire the real sequences.
Although the quality of each stage in our pipeline depends on the performance
of every other stage, i.e. a poor segmentation might be accountable for a
poor reconstruction, it is valuable to evaluate the different elements.

\noindent {\bf Pose estimation} We compared the estimated and ground-truth trajectories  by computing the absolute trajectory (AT)
root-mean-square errors (RMSE) for each of the objects in the
scene. Results on synthetic sequences are shown in
table~\ref{tab:at-rmse} and
Figure~\ref{fig:synth-errors}.  Results on the real GT sequences comparing estimated and GT  trajectories (given by OptiTrack) can be found in supplementary material \footnote{Please
see \url{http://visual.cs.ucl.ac.uk/pubs/cofusion/index.html} for
additional experimental evaluation and video.}.

\begin{table}[]
  \centering
\begin{tabular}{|ll|l|l|l|}
\hhline{-----}
%\tiny($>5cm$)
& Object & Error (avg/std, in mm) & Outlier-1cm & Outlier-5cm \\
\hhline{-----}
\multirow{3}{*}{\rowname{\vspace{-1mm} \scriptsize \emph{Esone1}}}
  & Head    & 3.216 / 5.94 & 4.38\%  & 0.016\% \\
  & Dice    & 5.805 / 7.27 & 19.86\% & 0.0\% \\
  & Gnome   & 5.051 / 6.10 & 12.39\% & 0.0\% \\
\hhline{-----}
\end{tabular}
\setlength{\belowcaptionskip}{-5pt}
\caption{Average error and standard deviation of the 3D reconstruction (mm) for the \emph{Esone1} ground truth scene (column 1). Percentage of surfels with reconstruction errors larger than $1cm$ (column 2) and $5cm$ (column 3).}
\label{tab:recon}
\end{table}

\begin{table}[]
  \centering
\begin{tabular}{|ll||l|l|l|}
\hhline{~~---}
\multicolumn{2}{c|}{} & Co-Fusion & ElasticFusion & Kintinuous \\
\hhline{--===}
\multirow{3}{*}{\rowname{\vspace{-1mm} \scriptsize \emph{ToyCar3}}}
  & Camera  & 6.126   & 5.917 & 0.999 \\
  & Car1    & 77.818  & -     & - \\
  & Car2    & 14.403  & -     & - \\
\hhline{-----}
\multirow{4}{*}{\rowname{\vspace{-1mm} \scriptsize \emph{Room4}}}
  & Camera        & 9.326             & 12.169  & 1.630 \\
  & Airship       & 9.108 / 10.118    & -       & - \\
  & Car           & 2.862             & -       & - \\
  & Rockinghorse  & 58.007            & -       & - \\
\hhline{-----}
\end{tabular}
\setlength{\belowcaptionskip}{-20pt}
\caption{AT-RMSEs of estimated trajectories for our synthetic sequences (mm).
Two trajectories are associated with the airship, since this object was split into two parts.}
\label{tab:at-rmse}
\end{table}

\noindent {\bf Motion segmentation} As the result of the segmentation
  stage is purely 2D, conventional metrics for segmentation quality
  can be used. We calculated the intersection-over-union measure per
  label for each frame of the synthetic
  sequences (we did not have ground truth segmentation for the real sequence). Figure~\ref{fig:synth-errors} shows the IoU for each
  frame in the \emph{ToyCar3} and \emph{Room4} sequences. %The overall
 % mean values \hl{...}

\begin{figure*}
  \def \imwidth {.26}
  \def \imheight {2.9cm}
	\centering
\subfloat[ATE (\emph{ToyCar3})]{\includegraphics[height=\imheight]{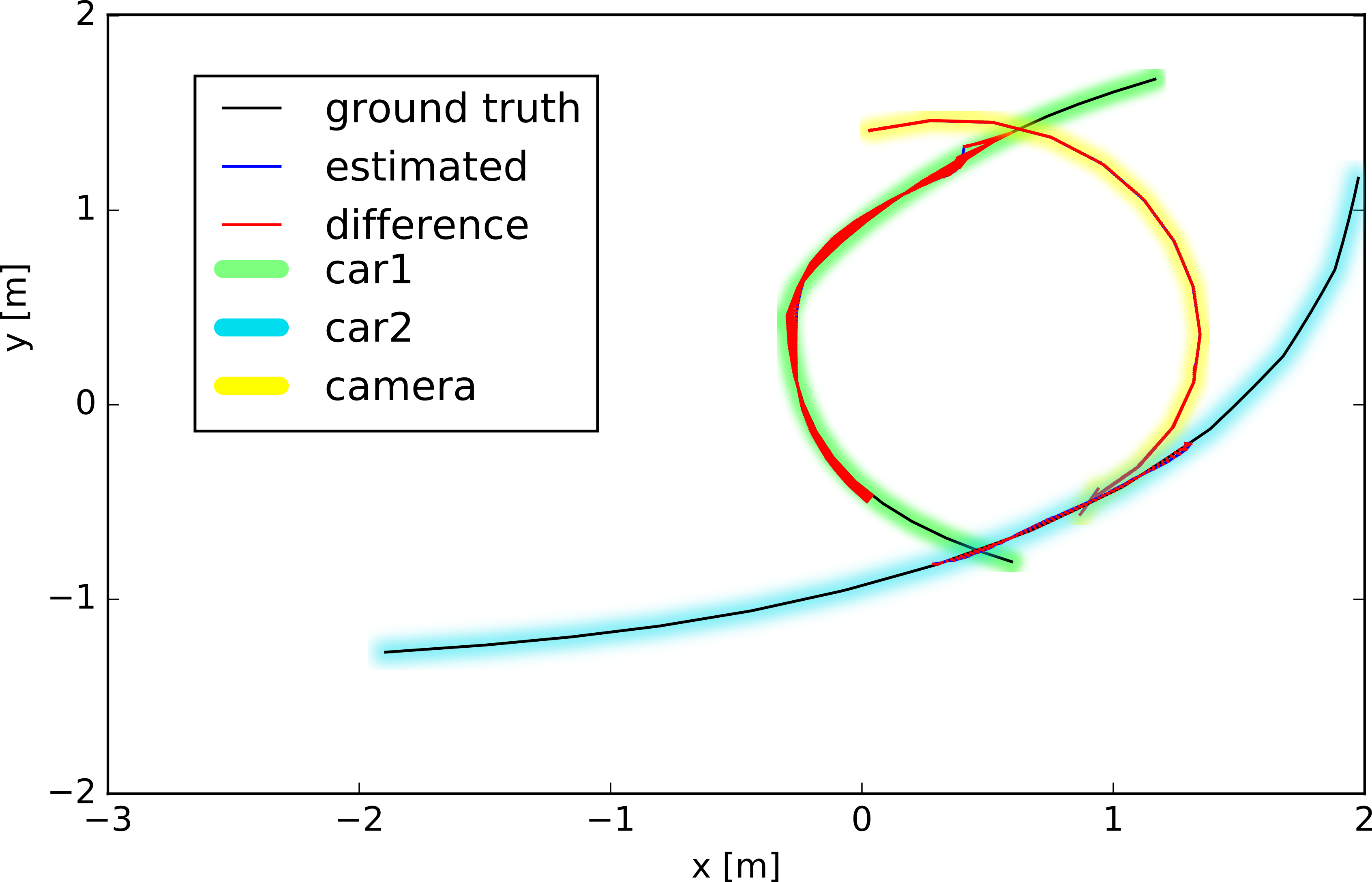}}~
\subfloat[ATE (\emph{Room4})]{\includegraphics[height=\imheight]{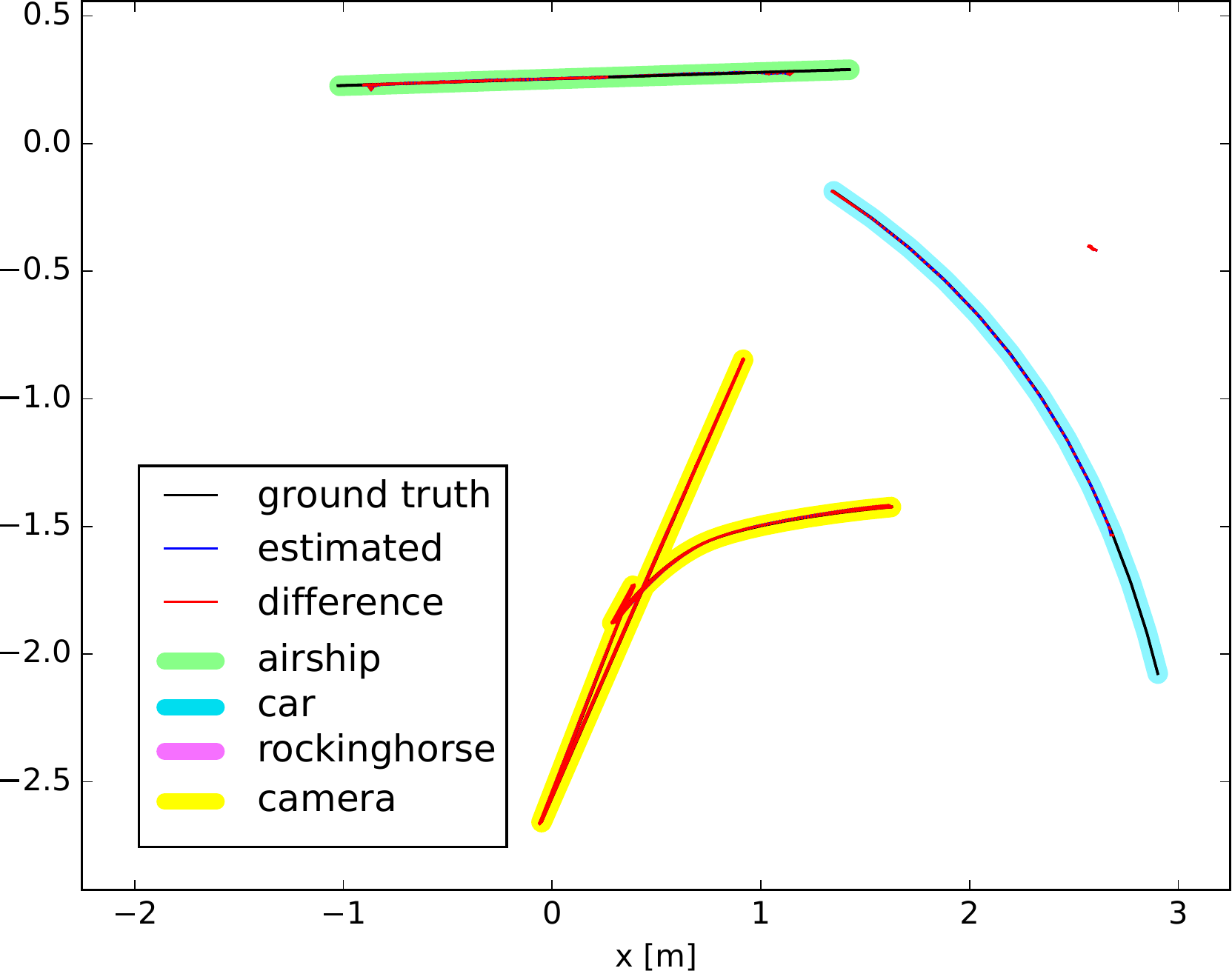}}
\subfloat[IoU (\emph{ToyCar3})]{\includegraphics[height=\imheight]{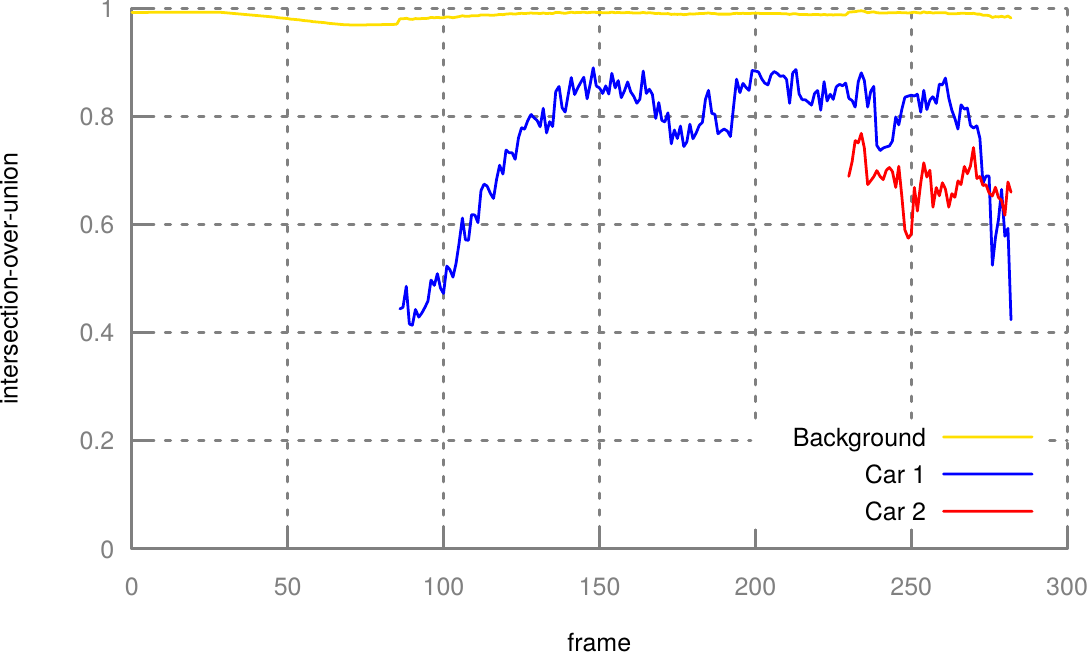}}
\subfloat[IoU (\emph{Room4})]{\includegraphics[height=\imheight]{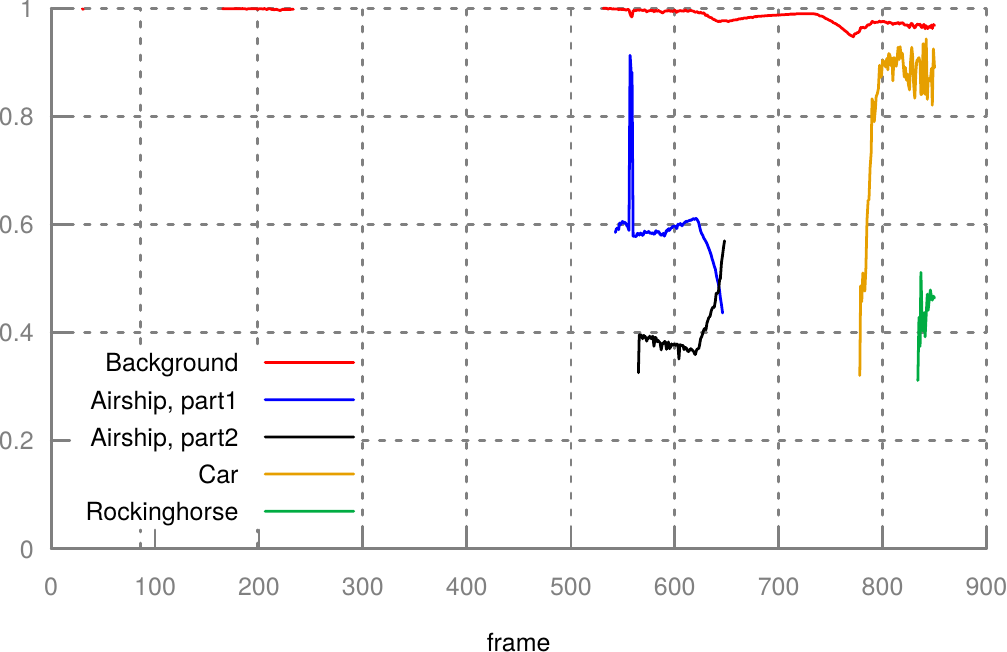}}
  \setlength{\belowcaptionskip}{-10pt}
	\caption{Comparison between the ground truth and estimated trajectories for each of the objects in the (a) \emph{ToyCar3} and (b) \emph{Room4} sequences. Intersection-over-union measure for each label and each frame in
  the  (c) \emph{ToyCar3} and (d) \emph{Room4} sequences. The graphs for car1 and car2 start to appear later
  in time, since the objects were not segmented before.}
	\label{fig:synth-errors}
\end{figure*}

\begin{figure}
	\centering
	\includegraphics[width=1\linewidth]{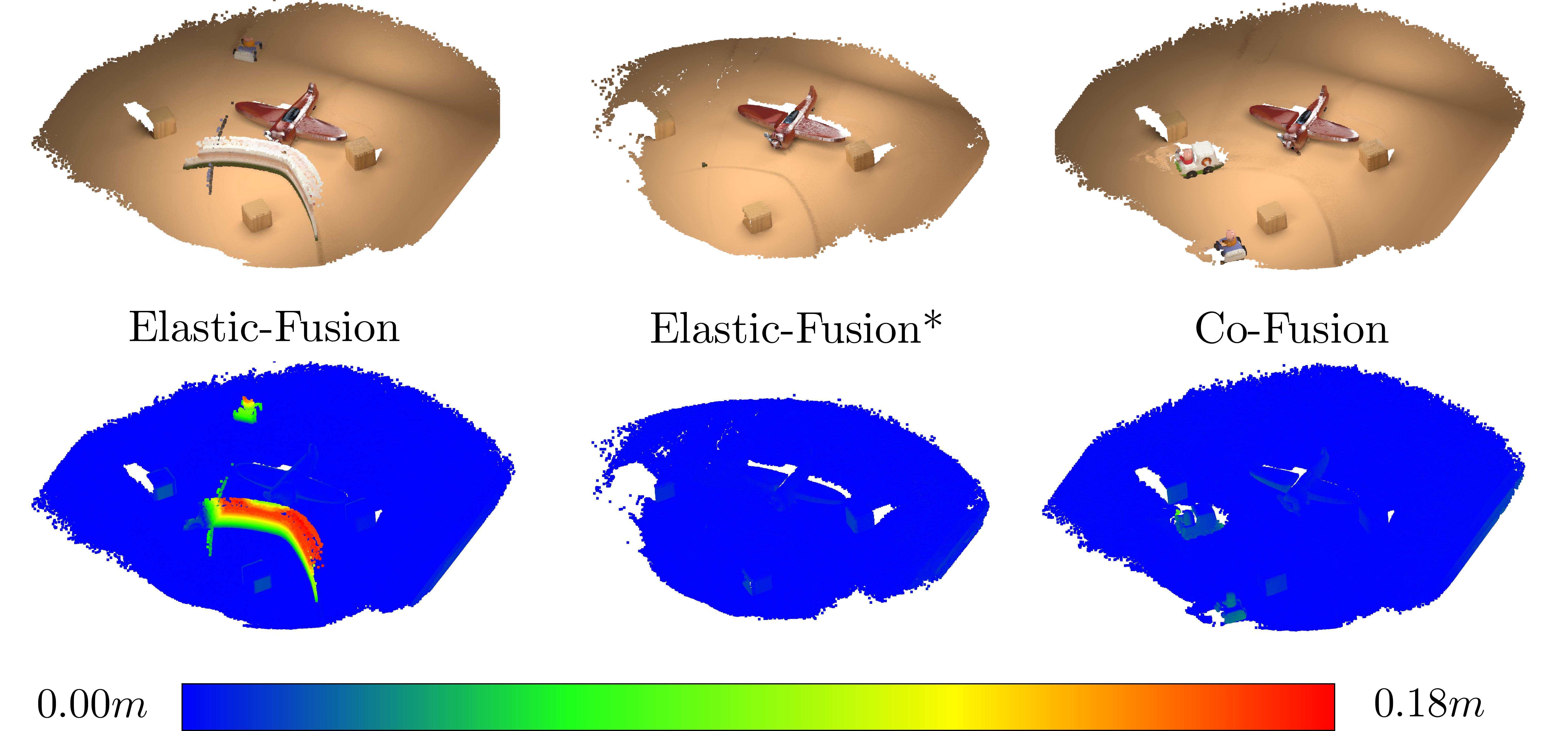}
  \setlength{\belowcaptionskip}{-20pt}
	\caption{This heat map compares the reconstruction error of Elastic-Fusion and Co-Fusion. As the original Elastic-Fusion implementation does not reject the geometry of moving objects completely, we added the outlier removal step described in Section~\ref{sec:fusion} for fair comparison (marked with *). Note that while the geometry of the toy cars is ignored by Elastic-Fusion*, it does appear in the reconstruction of Co-Fusion associated with low errors.}
	\label{fig:ef-mf-reconstruction}
\end{figure}

\noindent {\bf Fusion} To assess the quality of the fusion, one could either
inspect the 3D reconstruction errors of each object separately or jointly, by exporting
the geometry in a unified coordinate system. We used the latter on the synthetic sequences. This error is strongly conditioned on the tracking, but nicely highlights
the quality of the overall system. For each surfel in the unified map of
\emph{active} models, we compute the distance to the closest point on the
ground-truth meshes, after aligning the two representations. %\hl{Mean
                                                             %error
                                                             %values
                                                             %here}.
Figure~\ref{fig:ef-mf-reconstruction} visualizes the reconstruction error as a
heat-map and highlights differences to Elastic-Fusion. For the real scene \emph{Esone1}
we computed the 3D reconstruction errors of each object independently.
The results are shown in Table~\ref{tab:recon} and Figure~\ref{fig:esone}.

\begin{figure}
  \centering
  \subfloat[Front]{\includegraphics[width=0.4\linewidth]{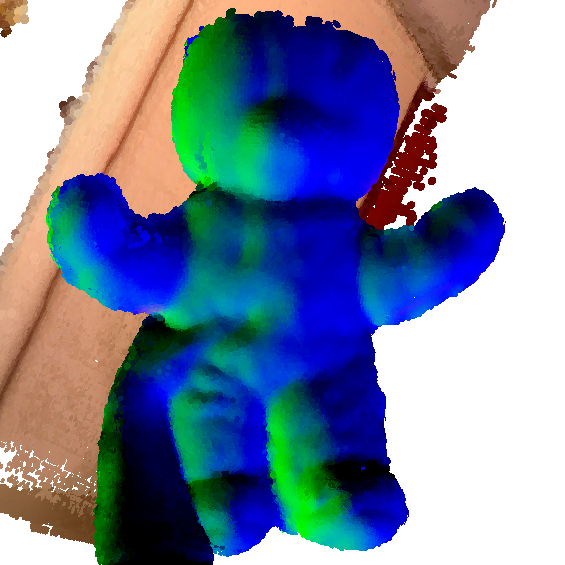}}~
  \subfloat[Back]{\includegraphics[width=0.4\linewidth]{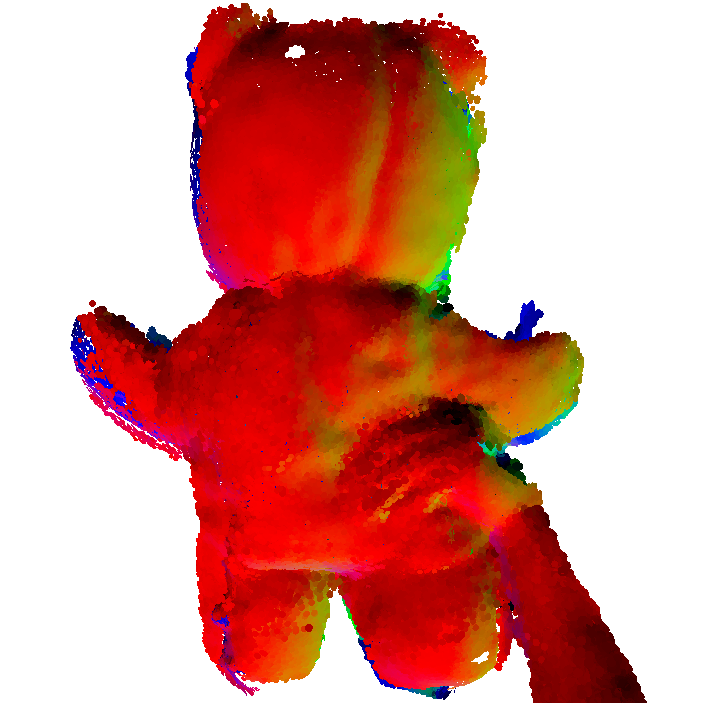}}
  \setlength{\belowcaptionskip}{-10pt}
  \caption{Hand-held reconstruction of a teddy bear: While the left hand was used rotate the teddy,
  the right one was holding the RGBD-sensor, which requires tracking of two independent motions.}
  \label{fig:teddy-reconstruction}
\end{figure}

\noindent {\bf Qualitative results} We performed a set of qualitative experiments to demonstrate the capabilities
of Co-Fusion.  One of its advantages is that it eases the 3D
scanning process, since we do not need to rely on the static-world
assumption. In particular, a user can hold and rotate an object in one
hand while using the other to move a depth-sensor around the object.
This mode of operation offers more flexibility, when compared to
methods that require a turntable, for
instance. Figure~\ref{fig:teddy-reconstruction} shows the result of
such an experiment.

\begin{figure}
  \def \imwidth {.16}
  \def \picwidth {0.1333} %(8/9) * \imwidth
  \def \xwidth {.2666} % 2 * \picwidth
  \center

  \settoheight{\tempdima}{\includegraphics[width=\imwidth\linewidth]{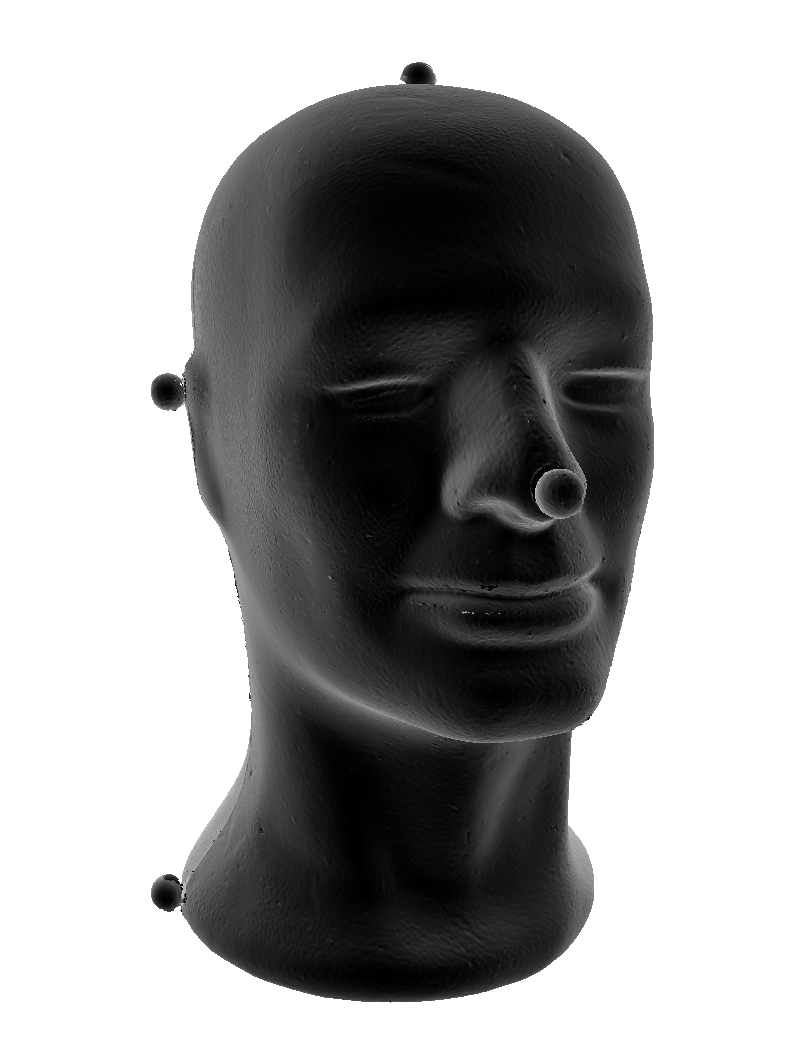}}
  \begin{tabular}{@{}c@{ }c@{ }c@{ }c|c@{ }c@{}}

  &Real&GT&Error&RGBD&\\
  \rowname{Head} &
  \includegraphics[width=\picwidth\linewidth]{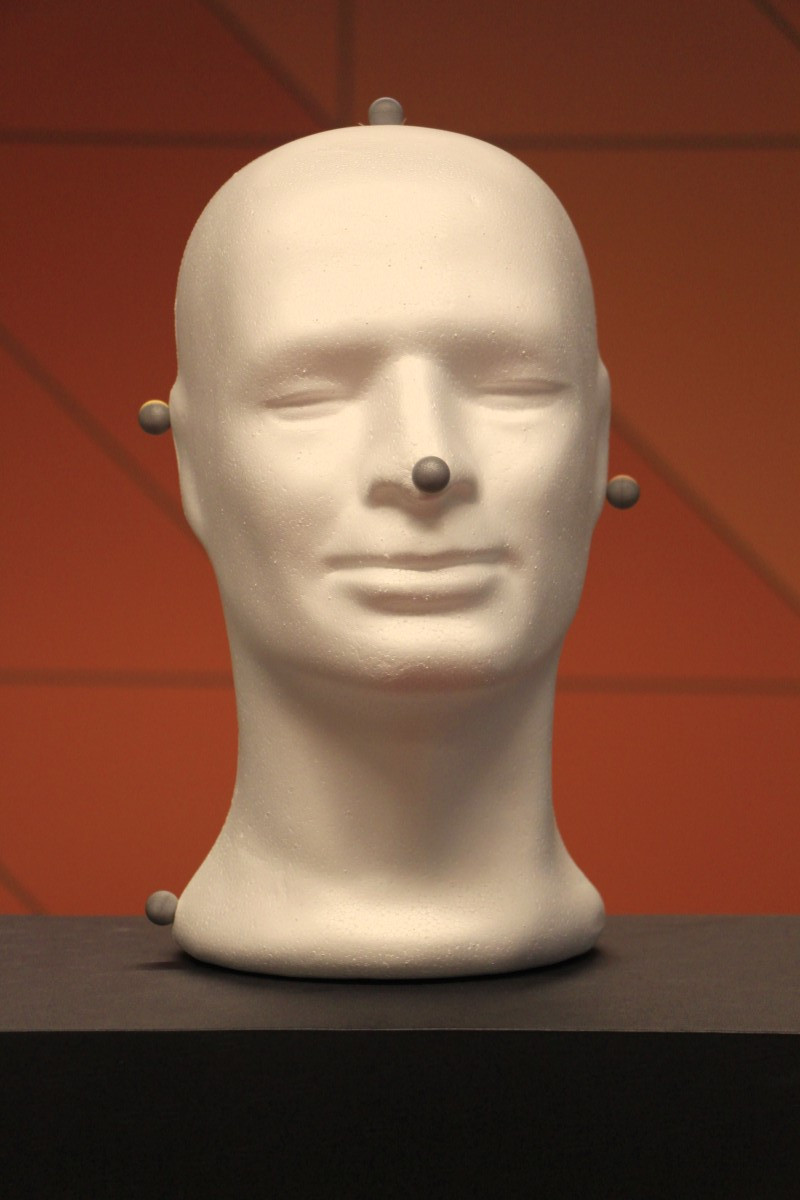} &
  \includegraphics[width=\imwidth\linewidth]{reconstruction-esone/head-rem34} &
  \includegraphics[width=\imwidth\linewidth]{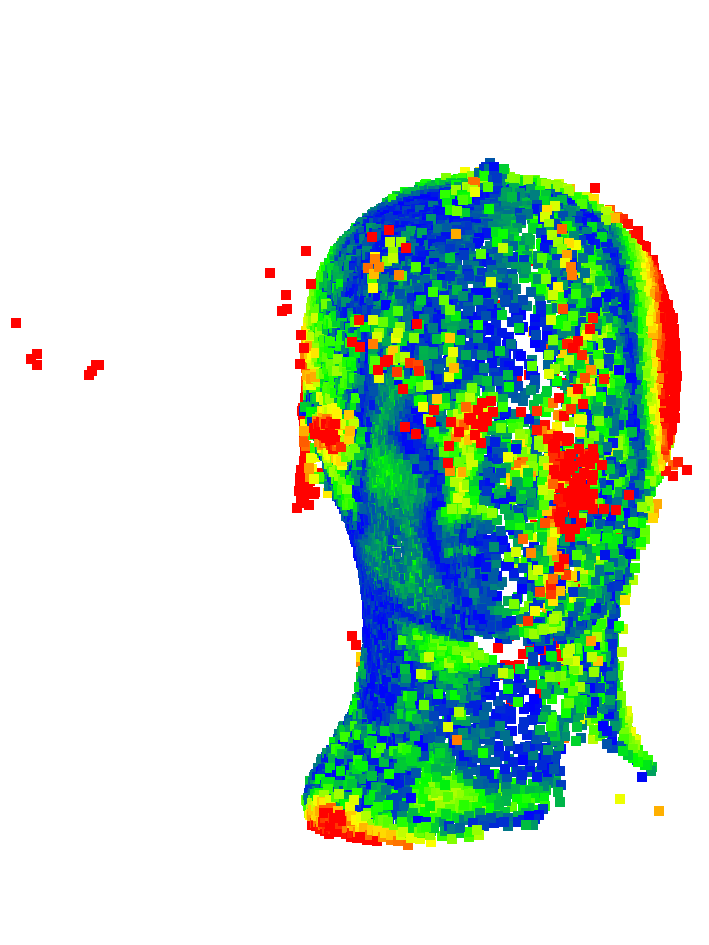} &
  \includegraphics[width=\xwidth\linewidth]{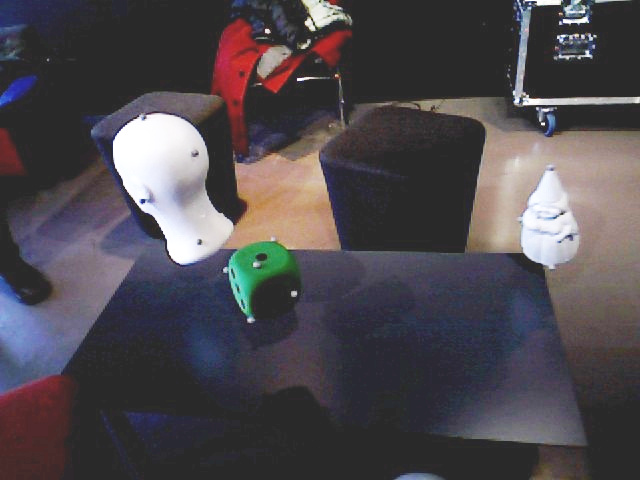} &
  \rowname{RGB} \\

  \rowname{Gnome} &
  \includegraphics[width=\picwidth\linewidth]{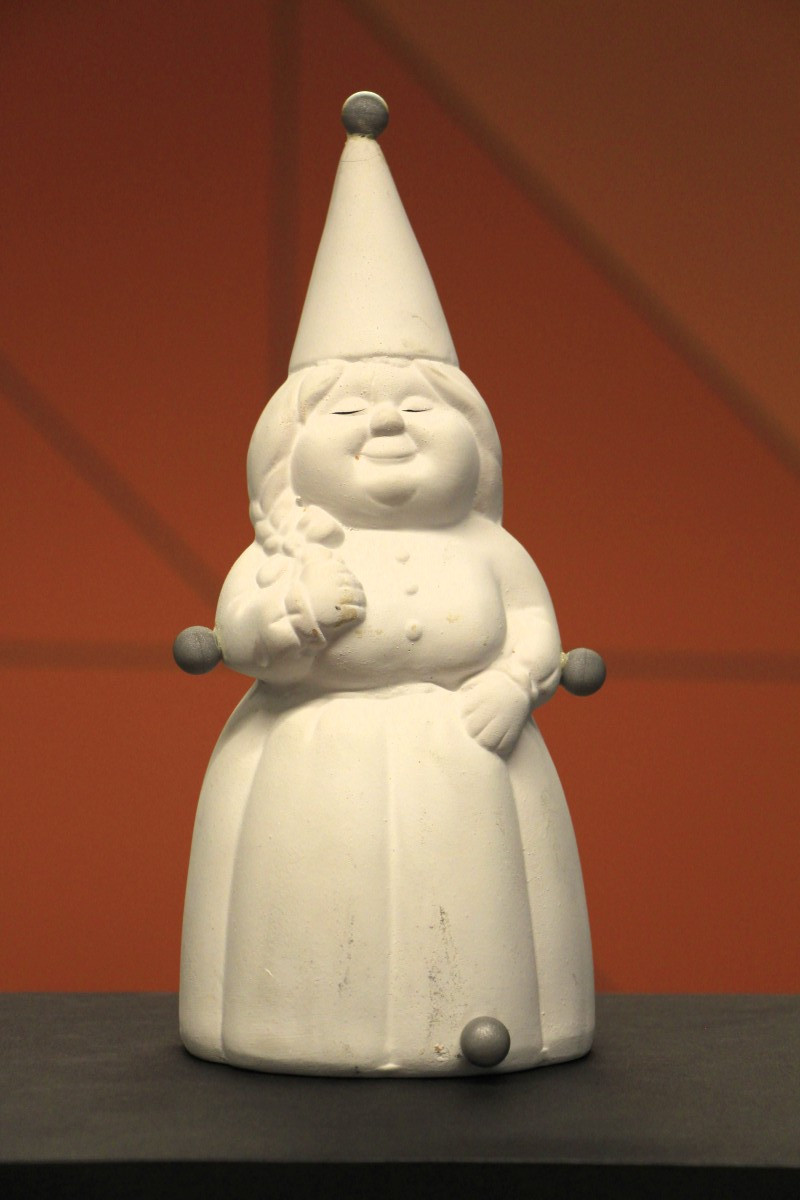} &
  \includegraphics[width=\imwidth\linewidth]{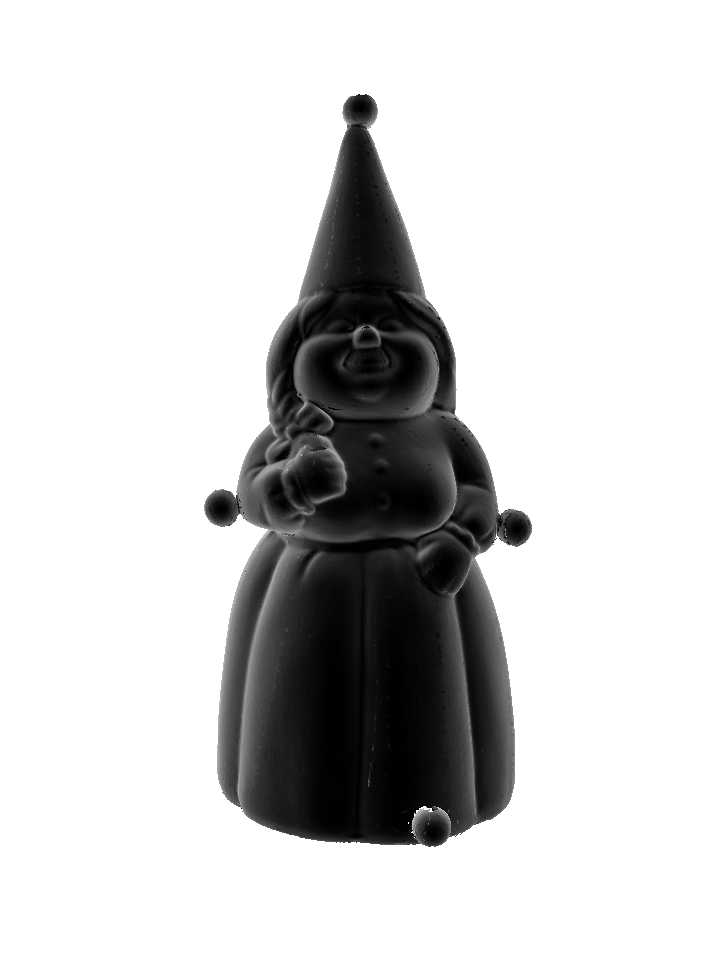} &
  \includegraphics[width=\imwidth\linewidth]{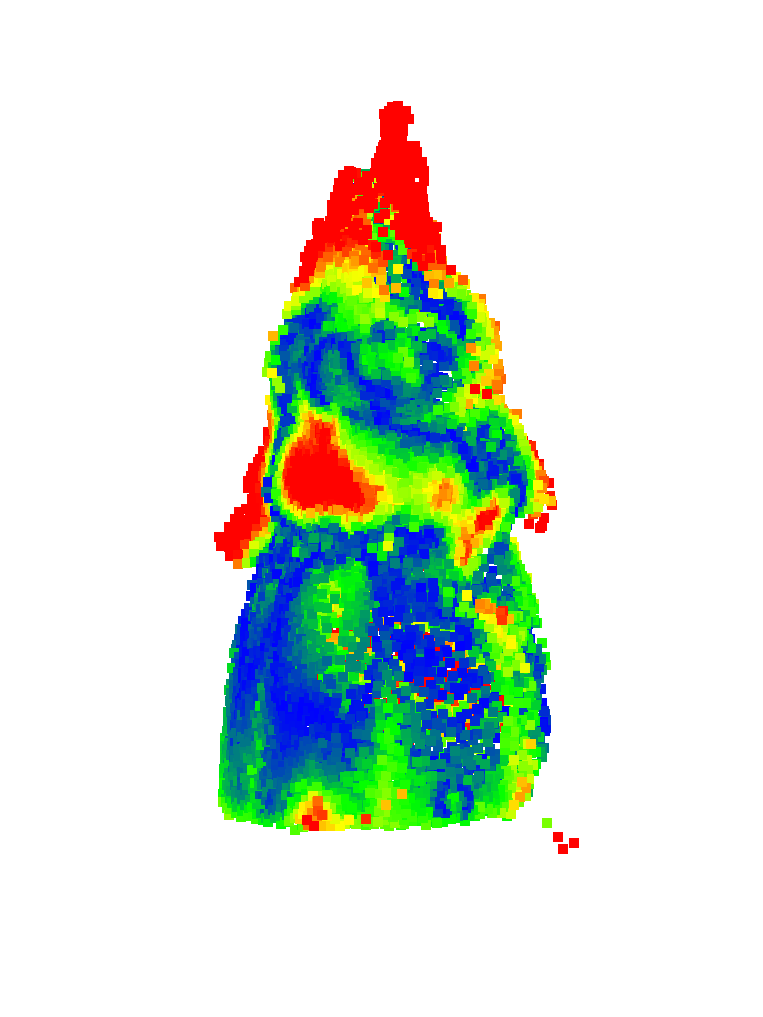} &
  \includegraphics[width=\xwidth\linewidth]{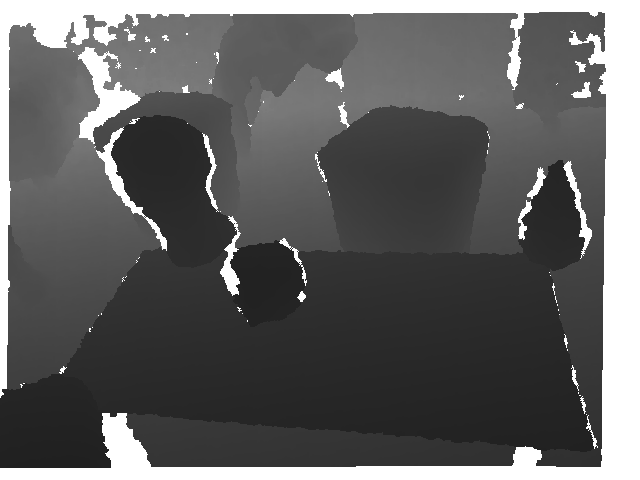} &
  \rowname{Depth} \\

  \rowname{Dice} &
  \includegraphics[width=\picwidth\linewidth]{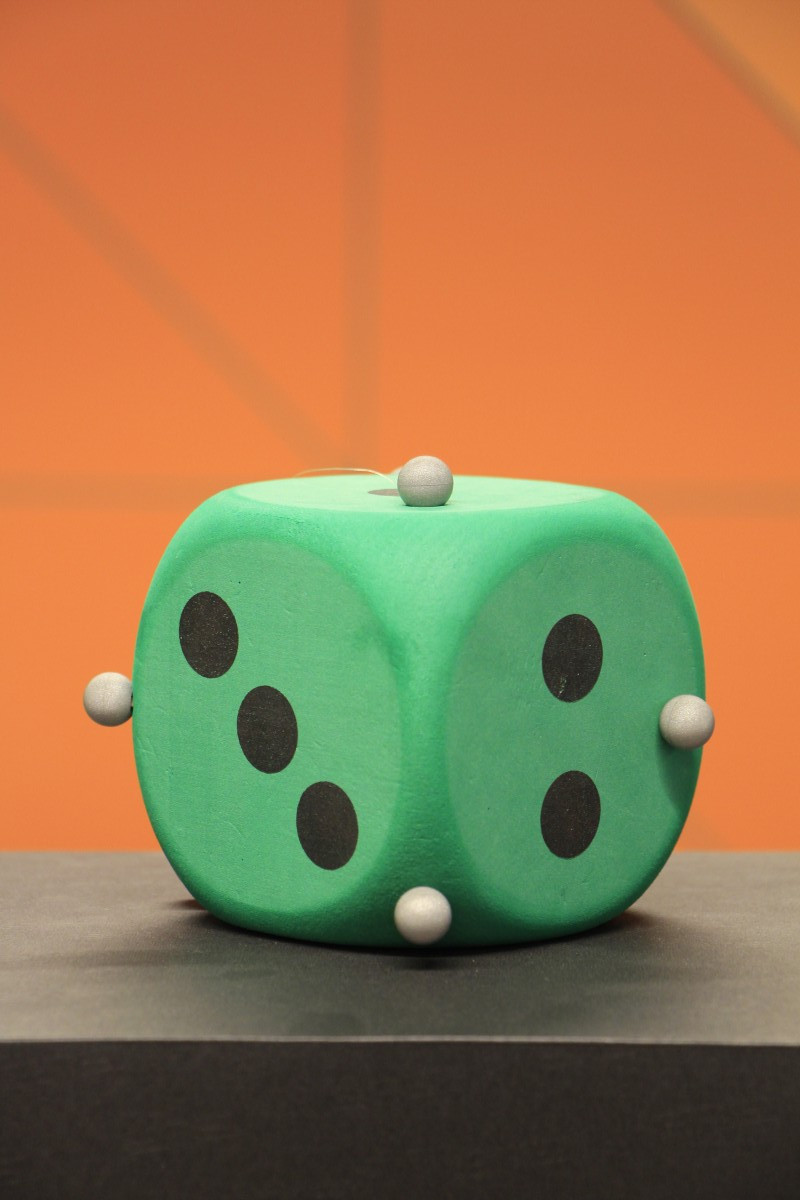} &
  \includegraphics[width=\imwidth\linewidth]{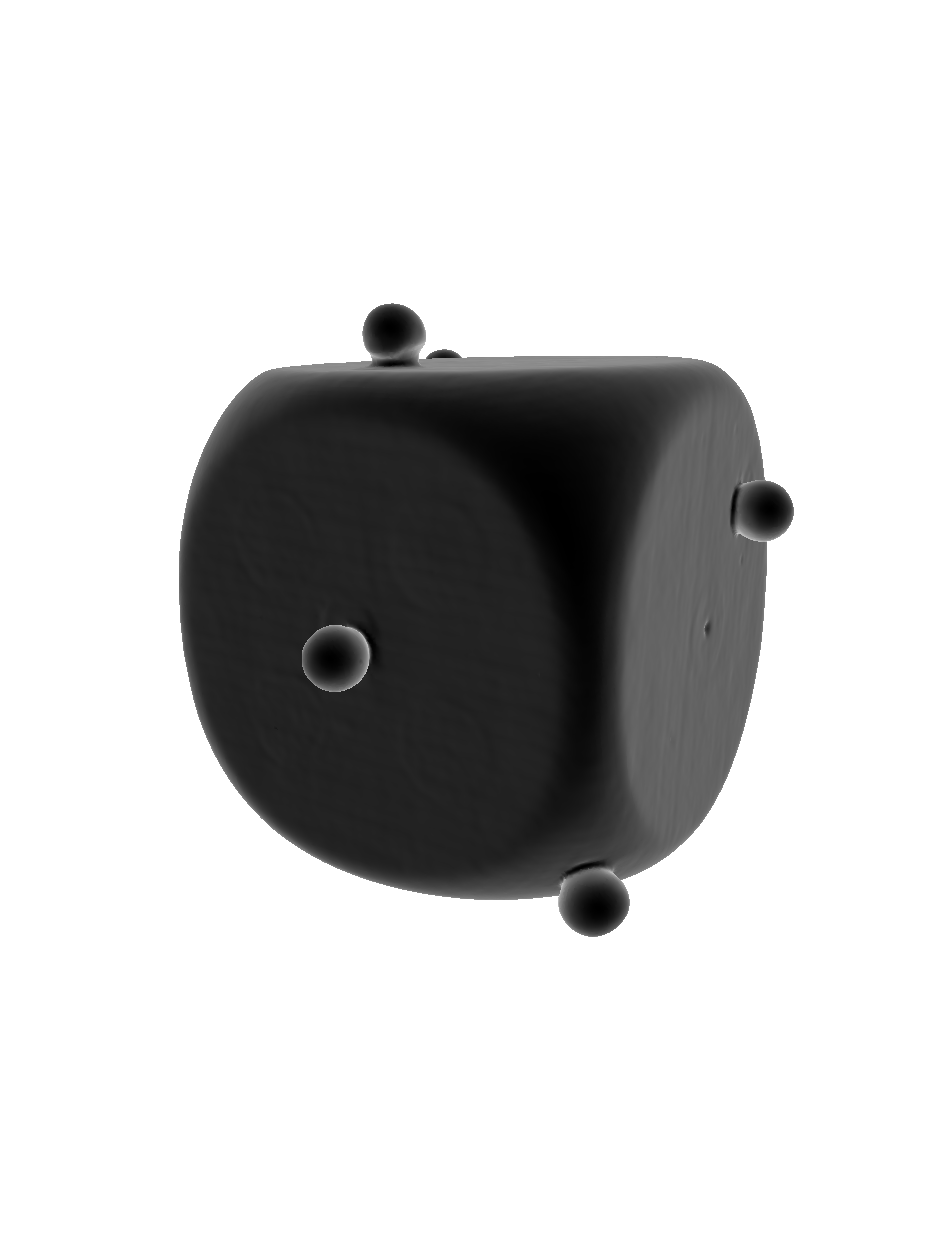} &
  \includegraphics[width=\imwidth\linewidth]{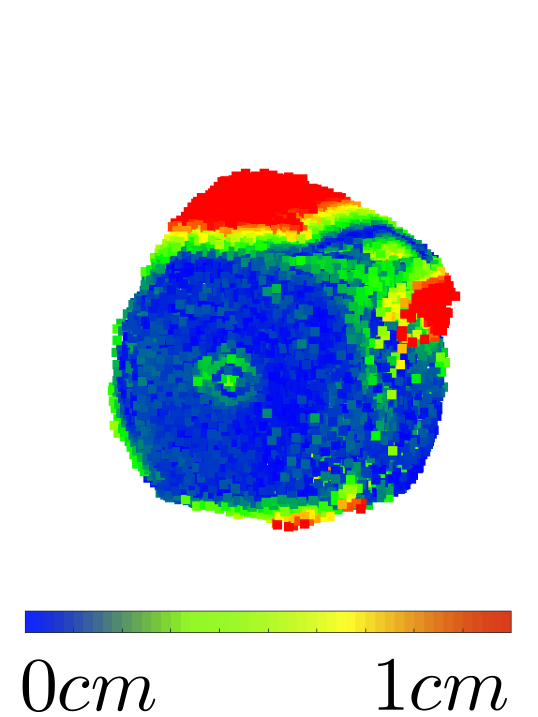} &
  \includegraphics[width=\xwidth\linewidth]{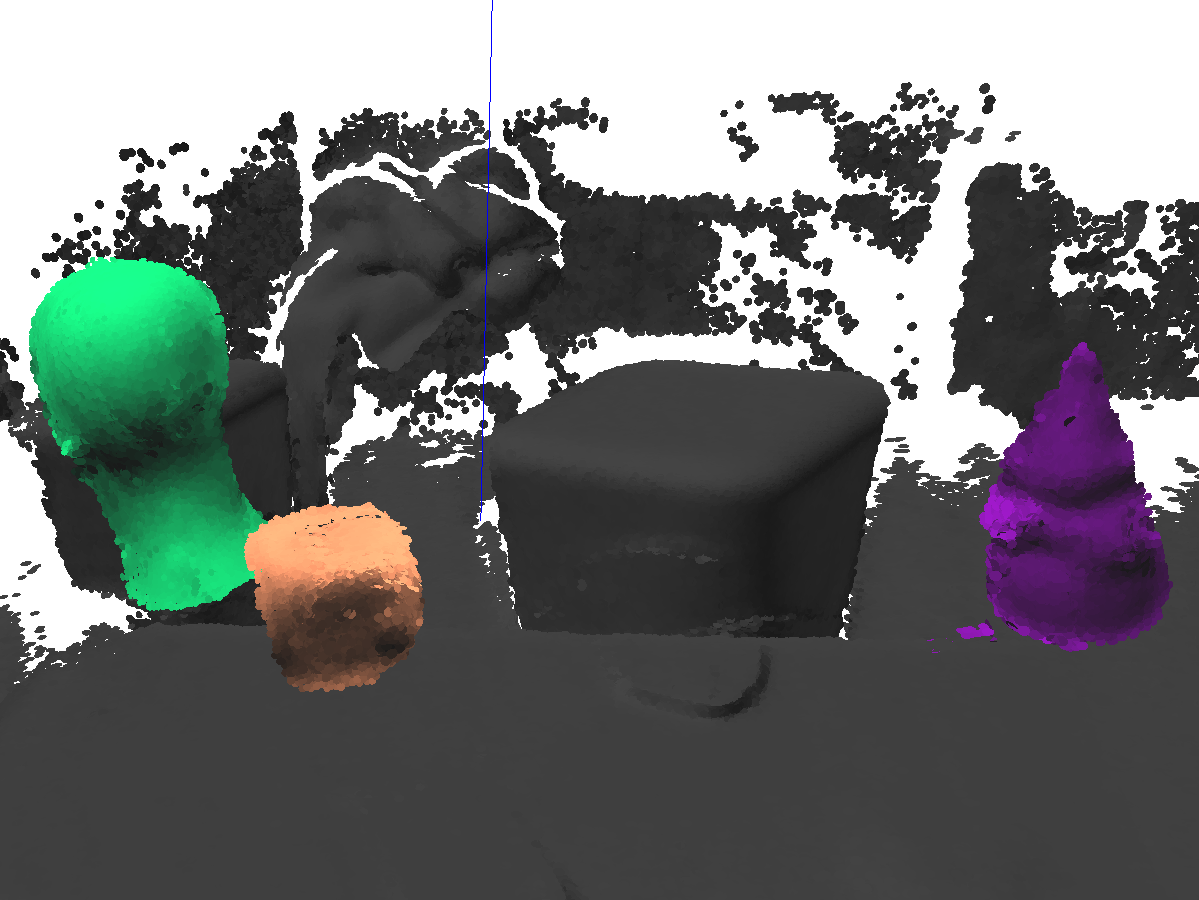} &
  \rowname{Labeling}

  \end{tabular}
  \setlength{\belowcaptionskip}{-10pt}
  \caption{Illustration of the \emph{Esone1} sequence. Markers were added to real 3D objects and tracked with an OptiTrack mocap system. A highly accurate 3D scanner was used to obtain ground-truth data of the geometry of the objects to allow a quantitative evaluation.}
  \label{fig:esone}
\end{figure}
Our final demonstration shows Co-Fusion continuously tracking and refining objects as they are placed on a table one after the other, as depicted in Figure~\ref{fig:placed-items}. This functionality can be useful in robotics applications, where objects have to be moved by an actuator. The result of the successful segmentation is shown in Figure~\ref{fig:placed-items}(b).

%%%%%%%%%%%%%%%%%%%%%%%%%%%%%%%%%%%%%%%%%%%%%%%%%%%%%%%%%%%%%%%%%%%%%%%%%%%%%%%%

\section{CONCLUSIONS}

We have presented Co-Fusion, a real time RGB-D SLAM system capable of
segmenting a scene into multiple objects using motion or semantic
cues, tracking and modeling them accurately while also maintaining a
model of the environment. We have demonstrated its use in robotics and
3D scanning applications.  The resulting system could enable a robot
to maintain a scene description at the object; even in
the case of dynamic scenes.

%%%%%%%%%%%%%%%%%%%%%%%%%%%%%%%%%%%%%%%%%%%%%%%%%%%%%%%%%%%%%%%%%%%%%%%%%%%%%%%%

\section*{ACKNOWLEDGMENT}
This work has been supported by the SeconHands project, funded from
the EU Horizon 2020 Research and Innovation programme under grant
agreement No 643950.

%%%%%%%%%%%%%%%%%%%%%%%%%%%%%%%%%%%%%%%%%%%%%%%%%%%%%%%%%%%%%%%%%%%%%%%%%%%%%%%%

\bibliographystyle{plain}
\bibliography{references}

\end{document}